\documentclass[10pt,twocolumn,letterpaper]{article}

\usepackage[pagenumbers]{cvpr} % To force page numbers, e.g. for an arXiv version

\definecolor{cvprblue}{rgb}{0.21,0.49,0.74}
\usepackage[pagebackref,breaklinks,colorlinks,allcolors=cvprblue]{hyperref}

\usepackage{amsmath}
\usepackage{booktabs}
\usepackage{graphicx}
\usepackage{algorithm}
\usepackage[noend]{algorithmic}
\usepackage{subcaption}

\usepackage{amsthm}
\theoremstyle{plain}
\newtheorem{theorem}{Theorem}[section]

\theoremstyle{definition}

\theoremstyle{remark}

\def\paperID{-----} % *** Enter the Paper ID here
\def\confName{CVPR}
\def\confYear{2026}

\title{Multi-Part Object Representations via Graph Structures and Co-Part Discovery}

\author{%
  Alex Foo \hspace{1.5em} Wynne Hsu \hspace{1.5em} Mong Li Lee\\
  National University of Singapore\\
  \texttt{\{dcsafdw,dcshsuw,dcsleeml\}@nus.edu.sg} \\
}

\begin{document}
\maketitle
\begin{abstract}
Discovering object-centric representations from images can significantly enhance the robustness, sample efficiency and generalizability of vision models.
Works on images with multi-part objects typically follow an implicit object representation approach, which fail to recognize these learned objects in occluded or out-of-distribution contexts. 
This is due to the assumption that object part-whole relations are implicitly encoded into the representations through indirect training objectives.
We address this limitation by proposing a novel method that leverages on explicit graph representations for parts and present a co-part object discovery algorithm.
We then introduce three benchmarks to evaluate the robustness of object-centric methods in recognizing multi-part objects within occluded and out-of-distribution settings.
Experimental results on simulated, realistic, and real-world images show marked improvements in the quality of discovered objects compared to state-of-the-art methods, as well as the accurate recognition of multi-part objects in occluded and out-of-distribution contexts.
We also show that the discovered object-centric representations can more accurately predict key object properties in a downstream task, highlighting the potential of our method to advance the field of object-centric representations.
\end{abstract}

\section{Introduction}

Human interaction with the environment relies on the comprehension of diverse visual scenes, powered by a robust visual system which decomposes each scene into abstract and manipulable objects \cite{kahneman1992reviewing,spelke2007core}.
Each object's features can be explicitly described, and different objects are recognized via the appearance and relative pose of their constituent parts \cite{hinton1979some}.
Emulating this part-whole compositional system in the form of multi-part object representations can improve robustness, occlusion-aware perception, generalization to out-of-distribution images, and interpretability of machine learning algorithms for vision \cite{greff2020binding}.
 
\begin{figure}[t!]
  \centering
  \includegraphics[width=1.0\linewidth]{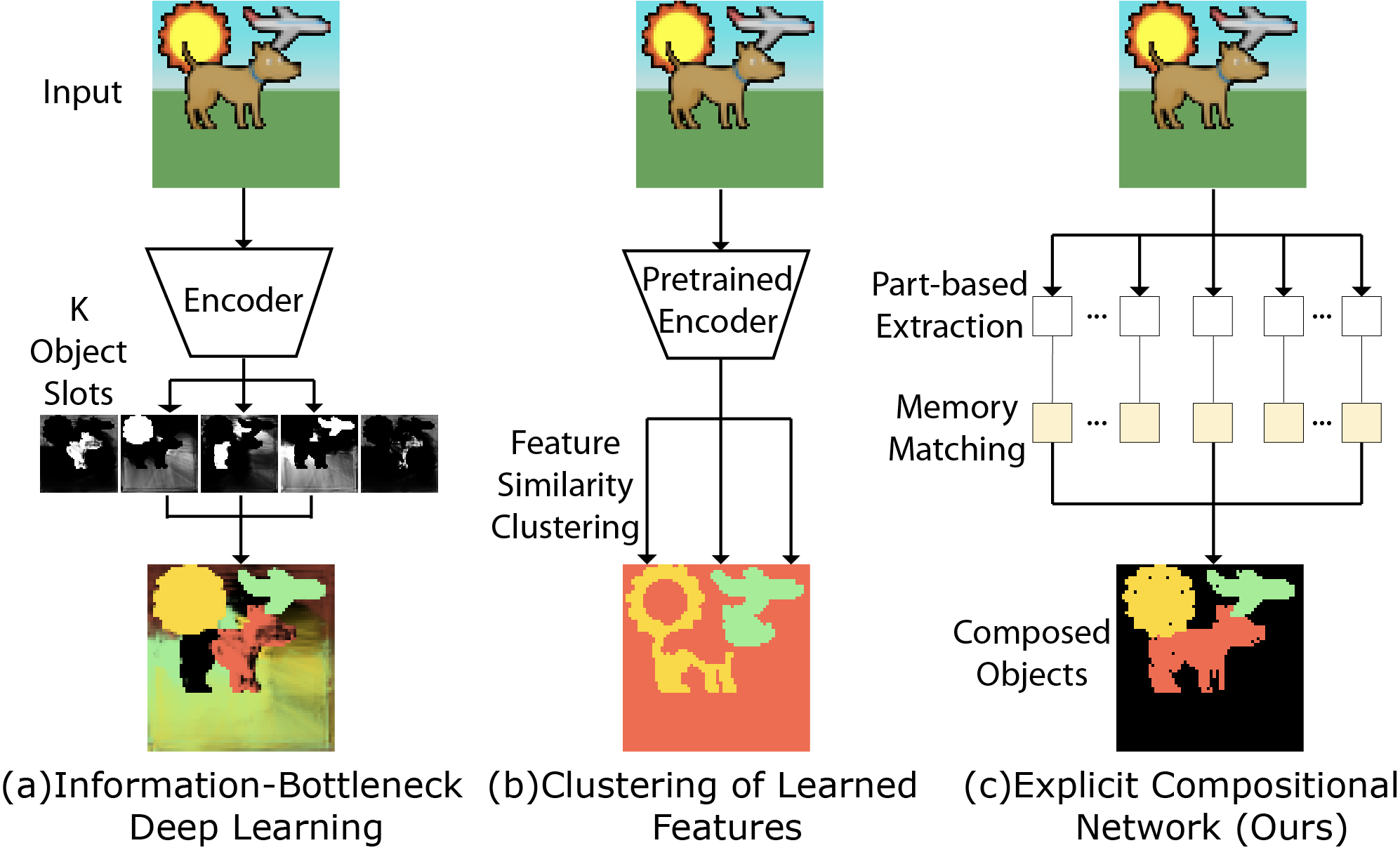}
   \caption{\small Comparison of existing methods. (a) Information-bottleneck approaches face challenges with multi-part object recognition, while (b) methods that rely on learned features struggle to compose part clusters into object wholes. Our proposed Explicit Compositional Network (ECO-Net) consistently and accurately composes parts into object wholes.}
   \label{fig:intro_comparisons}
\end{figure}

Recent work utilizes an implicit object representation approach in which scenes are decomposed either via: 
(a) information-bottleneck deep learning, where high-dimensional inputs are encoded into a fixed number of low-dimensional object representations then recombined for reconstruction \cite{
%crawford2019spatially, greff2019multi,LinWPSSDJA20, %bear2020learning, locatello2020object, EngelckeKJP20,
engelcke2021genesis,emami2021efficient, singh2021illiterate,singh2023neural,jia2023improving},
or
(b) similarity clustering of pre-trained image features \cite{simeoni2021localizing,seitzer2022bridging,wang2022self,wang2023cut,foo2023multi}.
Information-bottleneck approaches face challenges with multi-part object recognition, while methods that rely on learned features struggle to compose part clusters into object wholes (see Figure \ref{fig:intro_comparisons}).
The fundamental limitation of these methods stem from their assumption that object part-whole relations are implicitly encoded into the object representations through indirect training objectives such as input reconstruction \cite{yuan2022compositional} or self-supervised contrastive learning \cite{caron2021emerging}. 
While it is challenging to develop structures that consistently compose parts into their respective objects, 
this lack of part-whole structure in the unsupervised learning of objects leads to failure in recognizing objects in occluded and out-of-distribution contexts despite many training samples and iterations \cite{greff2020binding,DittadiPVSWL22, jiang2023object}.
These limitations impede deployment of these methods to real-world applications where robust recognition of multi-part objects is often essential \cite{yang2022promising}.

To address the limitations of implicit object representation,
we propose Explicit Compositional Network (ECO-Net), a framework that leverages on explicit graph representations for parts and learns to discover multi-part objects via a co-part object discovery algorithm.
Here, objects are defined as graphs, where 
the nodes define the features of each constituent part which make up the object, 
and the edges define the spatial relationships between the parts.
Based on the intuition neighboring parts that recurringly appear together across samples should belong to the same object \cite{brady2003bootstrapped,kruger2012deep}, 
we design a co-part object discovery algorithm which clusters parts together to form object wholes based on the repeated occurrence of parts and neighbors with similar positional relations.
Discovered objects are processed and integrated into an object memory module which is used for downstream tasks.
Figure \ref{fig:intro_comparisons} shows that our proposed Explicit Compositional Network (ECO-Net) effectively composes parts into object wholes.

Extensive experiments show that our framework outperforms state-of-the-art methods in discovering multi-part objects from simulated, realistic and real-world images in a fine-grained manner without supervision. 
We also introduce three benchmarks to evaluate the recognition of multi-part objects in occluded and out-of-distribution contexts. 
Our solution excels in accurately filling in the missing parts of objects in the occlusion context, while also accurately recognizing learned objects in the out-of-distribution context. 
Furthermore, our method predicts key object properties in a downstream task with greater precision.

Our contributions include: (1) a framework that utilizes explicit graph-based multi-part object representations, (2) a co-part object discovery algorithm as an alternative to indirect unsupervised learning objectives, (3) introduction of AbsScene-O, GSO-O and AbsScene-C benchmarks for evaluating robustness of object-centric methods on multi-part objects in occluded and out-of-distribution contexts, (4) experimental validation of our solution's superior performance on object discovery, occlusion-aware perception and out-of-distribution generalization, and (5) demonstration of the usefulness of learned object representations for predicting key object properties in a downstream task.

\section{Related Work}
Numerous works have shown remarkable success in segmenting real-world images, using supervised signals for
 image segmentation and object detection \cite{he2017mask, carion2020end, kirillov2023segment}. 
 Unsupervised methods like SLIC \cite{SLIC} employ a modified k-means algorithm to cluster pixels, 
while  Normalized Cut (Ncut) ~\cite{shi2000normalized} and Felzenszwalb's method \cite{felzenszwalb2004efficient} use
graph-based modeling on hand-crafted features for segmentation.
Others  use co-segmentation to jointly segment objects across images, but rely on intra-class objects being common to all images in the ground-truth \cite{
%rubinstein2013unsupervised,
li2019deep,choudhury2021unsupervised}. 
All these methods do not focus on obtaining useful object representations for the segmented components.
Unsupervised methods to obtain object-centric representations can be categorized as follows:

\textbf{$\bullet$ Information-bottleneck deep learning.} This approach uses two forms of attention to decompose scenes \cite{yuan2022compositional}.
Sequential attention methods uses RNN inspired models to  attend to different image regions.
MONet \cite{burgessMONetUnsupervisedScene2019} and GENESIS \cite{EngelckeKJP20} employ a deterministic network to perform the attention and sequentially discover and represent objects in the scene. 
These methods may neglect smaller objects as they tend to produce a weaker signal during the attention process, leading to incomplete or biased representations of scenes with objects of varying sizes.
To overcome this, GENESIS-V2 \cite{engelcke2021genesis} uses  stochastic stick-breaking  to perform attention randomly. 

Iterative attention methods randomly initialize a set of object representations  and  refine to bind these objects to different image regions. 
Slot Attention \cite{locatello2020object}  utilizes cross-attention along the object dimension. While it is fast and can be extended to handle videos \cite{kipf2022conditional}, it may fail to discover objects when the training set is diverse. 
BO-QSA \cite{jia2023improving} initializes the slots of Slot Attention as learnable embeddings and uses bi-level optimization, resulting in more stable training.
However, the number of clusters are fixed a priori which limits the applicability in real-world scenarios where the number of objects is not known beforehand.

\textbf{$\bullet$ Clustering of learned features.} This approach 
clusters similar regions into objects based on learned features.
DINOSAUR \cite{seitzer2022bridging} utilizes  self-supervised pre-training to obtain learned features before using iterative attention for feature reconstruction.
MaskCut \cite{wang2023cut} 
extracts image features using DINO and  applies Ncut for segmentation.
OC-Net \cite{foo2023multi} leverages learned feature connectivity to discover objects and designs two object-centric regularization terms.
The key limitation of feature-based clustering is the lack of modelling for complex part-whole relations of objects, which result in contrast-based segmentation of parts instead of composing parts into their proper objects.

\section{Proposed Approach}
\begin{figure*}[t!]
  \centering
  \includegraphics[width=0.94\linewidth]{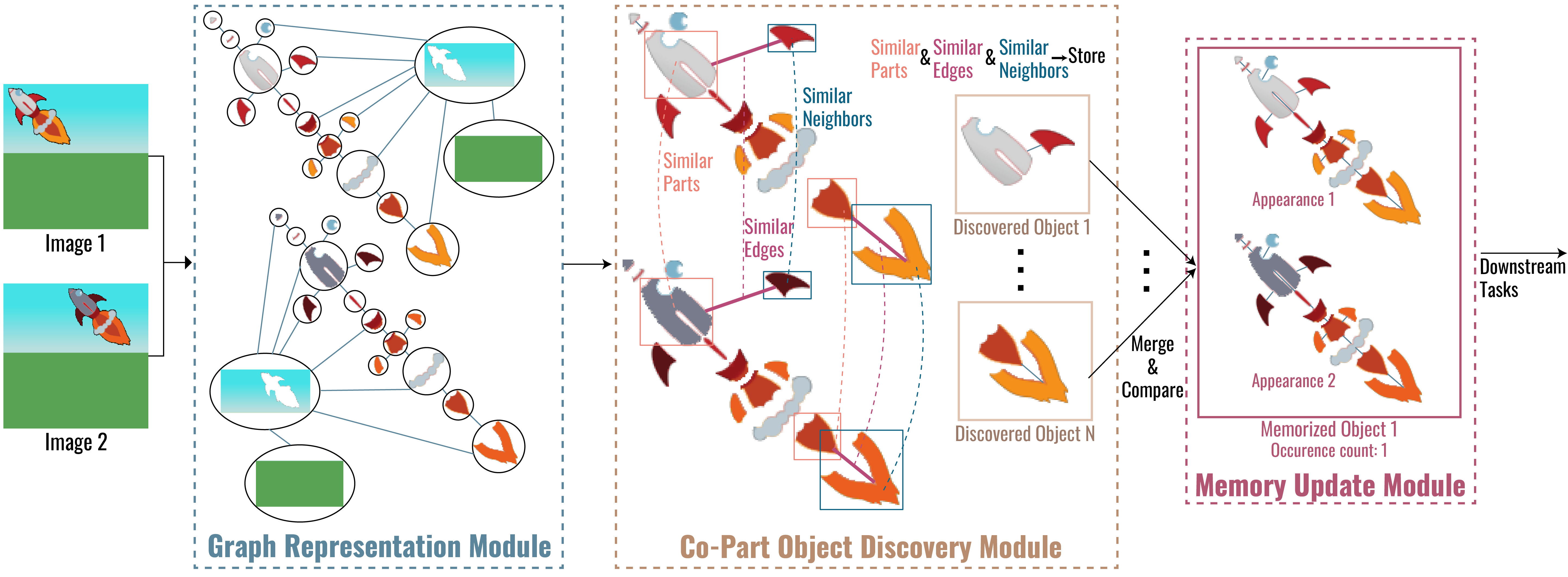}
   \caption{Overview of our proposed ECO-Net.}
   \label{fig:method_overview}
\end{figure*}
\begin{figure}[ht!]
  \centering
    \centering
    \includegraphics[width=0.67\linewidth]{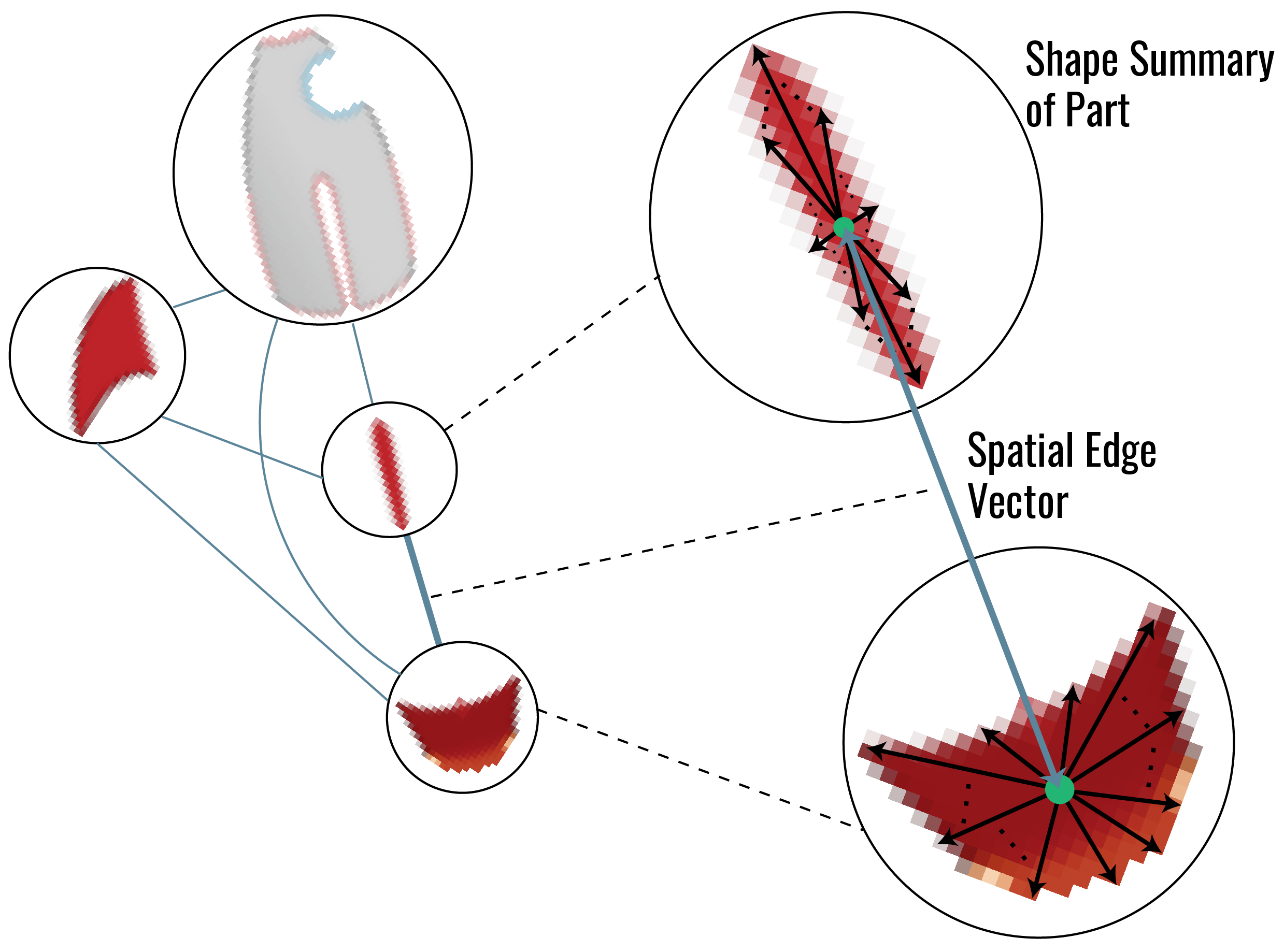} \\  (a) Graph of parts.\\
    
    \includegraphics[width=0.76\linewidth]{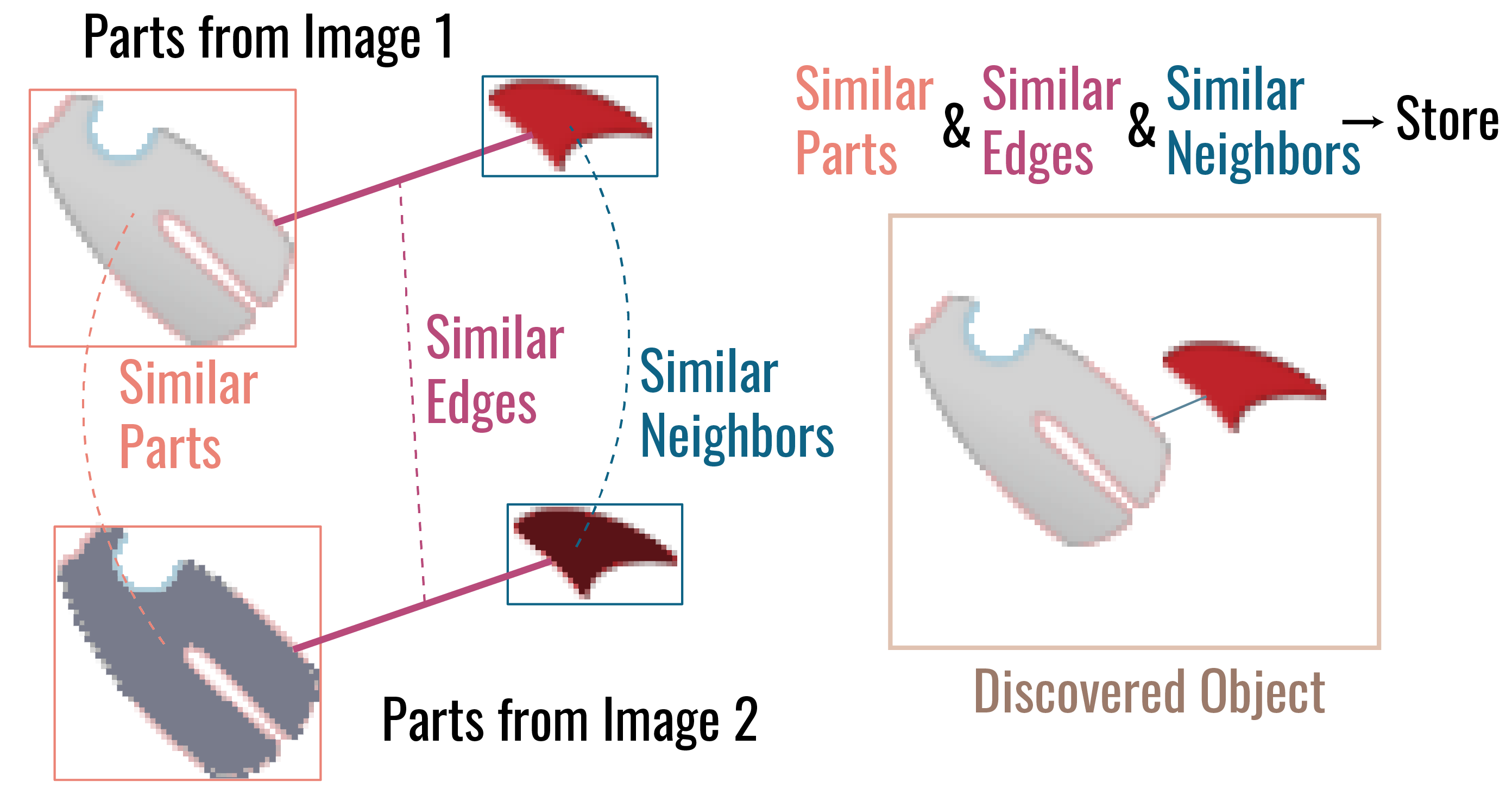}\\
    (b) Co-part object discovery.\\
    \label{fig:method_graph}

 \caption{Overview of object discovery. (a) Each node in the graph summarizes the shape of each part using vectors from its centroid to the sampled boundary pixels, while each edge depicts the spatial relationships between parts. (b) Repeated occurrence of parts with similar spatial relationships are grouped into objects.}
 \label{fig:objectdiscovery}
%  \end{minipage}
% \vspace{-10pt}
\end{figure}

Our approach extracts multi-part objects in images without relying on labeled data nor the need to specify the number of objects in the image. 
This is achieved by discovering part-whole relations via the recurring appearance of similarly related parts and neighbors across images. 
By not requiring the number of objects to be specified beforehand and accounting for part-whole relations, we can generalize to scenes with varying numbers of multi-part objects in real-world, occluded and out-of-distribution environments.

Figure \ref{fig:method_overview} shows an overview of  ECO-Net with three modules: (a) graph representation module that takes an image batch as input and use Felzenszwalb's algorithm to discover fine-grained parts in each image;
(b) co-part object discovery module clusters parts and their corresponding relations that repeatedly occur into objects; and (c) 
memory update module that processes the current set of objects discovered
and integrates it with the objects in memory. 

\paragraph{Graph Representation.} Felzenszwalb's  algorithm has theoretical guarantees in producing segments that obey global properties.
It ensures that  all segments adhere to the boundary predicate, and are invariant to the order in which the segmentation is performed. 
Given an input batch with images of $N$ pixels whose coordinates are $\mathcal{X} = \{\mathbf{x}_1,\dots,\mathbf{x}_N\}$, 
it outputs a batch-wise assignment of these pixels into disjoint sets which we call parts $\mathcal{P} = \{P_1,\dots,P_M\}$.
We  define a graph of parts $\hat{\mathcal{G}} = (\mathcal{P}, \hat{\mathcal{E}})$ where
 each node summarizes the position-invariant shape of each part, and
 each edge has a vector that depicts the spatial relationship between parts (see Figure \ref{fig:objectdiscovery}a).
For each part $P_i \in \mathcal{P}$,
we  sample $K$ boundary pixels  with coordinates $\{\mathbf{x}^i_1,\dots,\mathbf{x}^i_K\}$. 
Let $\mathbf{x}^i_c$ be the coordinates of the centroid of $P_i$. Then the shape of $P$ can be described in a position-invariant manner using $K$ vectors that point from the centroid to each boundary pixel 
$ \mathbf{V}_i = \big[\mathbf{x}^i_1 - \mathbf{x}^i_c,\cdots,\mathbf{x}^i_K- \mathbf{x}^i_c\big]$.
An edge $(P_i,P_j)$ exists if $P_i$ is in the same image as $P_j$ and the label of this edge is a vector $\mathbf{x}_{c}^j - \mathbf{x}_c^{i}$ that describes the spatial relationship of $P_i$ and $P_j$.

\paragraph{Co-Part Object Discovery.}
This module finds repeated occurrences of parts with similar spatial relationships across the image batch containing inter-class multi-part objects and groups them into objects (see Figure \ref{fig:objectdiscovery}b).
Given the graph of parts $\hat{\mathcal{G}} = (\mathcal{P}, \hat{\mathcal{E}})$ from an input batch of images, a solution to the general problem of composing recurring parts into their respective objects requires checking all possible subgraphs. 
Unfortunately, there are $2^{|\hat{\mathcal{E}}|}$ possible subgraphs, so enumerating them is infeasible. This problem requires repeatedly solving the subgraph isomorphism problem as a subroutine, which is NP-complete and intractable \cite{kuramochi2001frequent}. To overcome this, we leverage the intuition that only neighboring parts that recur together across multiple samples are likely to belong to the same object. 
Theorem \ref{theo:consubgraph} formalizes this, with its proof provided in Appendix \ref{appn:copart}.

\begin{theorem}
\label{theo:consubgraph}
Given a graph of parts $\hat{\mathcal{G}} = (\mathcal{P}, \hat{\mathcal{E}})$ derived from a batch of images, there exists a pairwise iterative clustering algorithm that solves the composition of recurrent parts with similar spatial relations into objects in $O(|\mathcal{P}|^2 \log |\mathcal{P}|)$ time, assuming each object is represented by a connected subgraph of neighboring parts.
\end{theorem}

Following Theorem \ref{theo:consubgraph}, we design a clustering algorithm that takes as input the graph of adjacent parts $\mathcal{G}=(\mathcal{P},\mathcal{E})$, where $\mathcal{E}\subset{\hat{\mathcal{E}}}$, and output the set of objects (see Appendix \ref{co-part-objdisc}).
We uniformly sample an unprocessed part $P_i$ and find parts that are similar based on their feature vectors.
The similarity between a pair of parts is the average cosine similarity  between their corresponding node features:
%\begin{equation} 
% \hspace*{0.3in}
\begin{equation*}
    \text{sim}(\mathbf{V}_{i}, \mathbf{V}_{j}) = \frac{1}{K}\sum_{k=1}^{K}\frac{(\mathbf{x}^{i}_k- \mathbf{x}^{i}_c) \cdot (\mathbf{x}^{j}_k- \mathbf{x}^{j}_c)}{\parallel(\mathbf{x}^{i}_k- \mathbf{x}^{i}_c)\parallel\parallel(\mathbf{x}^{j}_k- \mathbf{x}^{j}_c)\parallel}
\end{equation*}

For each part $P_j$ similar to $P_i$ with value more than $\epsilon$, 
let $\mathcal{E}_i$ and $\mathcal{E}_j$ be the set of edges connected to parts $P_i$ and $P_j$ respectively. 
For pairs of cosine-similar edges $(P_i,P_{k})\in \mathcal{E}_i$ and $(P_j,P_{k'}) \in \mathcal{E}_j$,
we consider that an object relation has been discovered if the corresponding neighbors $P_k$ and $P_{k'}$ are similar.
We express the discovered object as a combined set of its parts $\{\mathbf{V}_i,\mathbf{V}_k\} \in \mathcal{P}_c$ and connecting edge $\mathcal{E}_c$.
If the parts $\mathbf{V}_i,\mathbf{V}_k$ have not been assigned to some object, we add the new object into the set of objects found thus far $\mathcal{O}$.
If either $\mathbf{V}_i$ or $\mathbf{V}_k$ already belong to some other object, we extend the first existing object by appending the new parts and connecting edge, update edges with the new part, and merge all other existing objects with $\mathbf{V}_i$ or $\mathbf{V}_k$ into the first.
The process is repeated till all parts have been processed.

\paragraph{Memory Update.} This module integrates the set of objects discovered in the current batch with those stored in memory. For each discovered object, we first take the union of all pixels from the constituent parts and obtain a position-invariant representation of the object shape by sampling $K$ vectors which point from the centroid pixel to the boundary pixels. Variations in object appearance and pose is also summarized into this object representation as different views of the same object.
We then verify if the object exists in memory by comparing the similarity of its feature vectors against the feature vectors of the objects in memory. 
If a match is found, we increment the occurrence count. Otherwise, we add this new object to memory. 

Once all training samples have been processed, the object memory is sorted based on the number of occurrences to prioritize frequently occurring objects. This can then be utilized for downstream object discovery tasks. 

\section{Performance Study}
\begin{table*}[t!]
   \caption{Summary of dataset characteristics}
   \vspace*{-0.15in}
   \small
	\begin{center}
		\label{tab:datasets}
		\begin{tabular}{ccccccc}
			\toprule
			Dataset & Type & Ground Truth & Image Size & \# Samples & Min Visible \\
            \midrule
            Tetrominoes & Simulated & Pixel Mask & $35\times 35$ & 63K & 100\% \\
            AbsScene & Simulated & Pixel Mask & $64\times 64$ & 63K & 100\% \\
            GSO & Realistic & Pixel Mask & $128 \times 128$ & 63K & 100\%\\
            SKU-110K & Realistic & Bounding Box & $128 \times 128$ & 11K & Varied \\
            PASCAL VOC & Real-World & Pixel Mask & $128 \times 128$ & 12K & Varied \\
            MS COCO & Real-World & Pixel Mask & $128 \times 128$ & 12K & Varied \\
            AbsScene-O & Occluded Objects & Pixel Mask & $64\times 64$ & 63K & 25\% \\
            GSO-O & Occluded Objects & Pixel Mask & $128 \times 128$ & 63K & 25\%\\
            AbsScene-C & Out-of-Distribution & Pixel Mask & $64\times 64$ & 1K & 100\% \\
            \bottomrule
		\end{tabular}
	\end{center}
\end{table*}

We  evaluate the performance of ECO-Net in terms of quality, occlusion-aware perception, generalizability and downstream property prediction on a diverse range of datasets:

$\bullet$ \textbf{Simulated datasets} Tetrominoes \cite{multiobjectdatasets19} with Tetris-like objects sampled from 6 colors and 17 shapes, and AbsScene \cite{zitnick2013bringing} which has 10 multi-part objects.

$\bullet$  \textbf{Realistic datasets} GSO \cite{downs2022google} comprises photo-realistic images of common objects with varied lighting and 3D object positions, and SKU-110K \cite{goldman2019precise} is a challenging dataset with 1.7 million objects in dense retail scenes of various scales, angles and lighting. 

$\bullet$ \textbf{ Real-world multi-object datasets} PASCAL VOC \cite{pascal-voc-2012} is the widely used benchmark for object segmentation with annotations for 20 object categories.   
MS COCO is the variant of the Microsoft Common Objects in Context dataset \cite{lin2014microsoft, yang2022promising}.
    
    $\bullet$ \textbf{Occluded objects datasets} AbsScene-O and GSO-O. We generate these occlusion-heavy variations from the original AbsScene and GSO datasets, where occlusion of objects is allowed up to $25\%$ minimum visibility.
    
    $\bullet$ \textbf{Out-of-distribution dataset} AbsScene-C. We generate this test set by  randomly replacing the original background with 15 real-world complex-textured backgrounds.

Table \ref{tab:datasets} shows  the dataset characteristics (Details  in Appendix \ref{appn:datasets}). Following \cite{goldman2019precise}, we split SKU-110K into 8233 images for training and 588 images for testing. 
For PASCAL VOC, we use the augmented training set with 10582 samples and 1499 validation samples for testing. 
For MS COCO, we use the first 10K samples for training and the last 2K samples for testing. 
For the remaining datasets, we follow \cite{locatello2020object,engelcke2021genesis} and use the first 60K samples for training and the next 320 samples for testing.

\noindent\textbf{Baselines.} We compare ECO-Net with SLIC \cite{SLIC}, Ncut \cite{shi2000normalized}, Felzenszwalb \cite{felzenszwalb2004efficient}, Slot Attention \cite{locatello2020object}, GENESIS-V2 \cite{engelcke2021genesis}, BO-QSA \cite{jia2023improving}, DINOSAUR \cite{seitzer2022bridging}, MaskCut \cite{wang2023cut} and OC-Net \cite{foo2023multi}. Appendix \ref{appn:baselines} gives the details of these baselines. 

We train ECO-Net for 10K iterations with a batch size of 2, and set $\epsilon = 0.99$.
Training on AbsScene on a single GPU with 32GB of RAM takes 73 minutes.
For all methods, we set the number of foreground objects to 4 for Tetrominoes, 5 for AbsScene, 8 for GSO, 50 for SKU-110K, 6 for PASCAL VOC and MS COCO. 
We train information-bottleneck methods for 300K iterations with a batch size of 64, using Adam optimizer with a learning rate of $4\times10^{-4}$. 
We set the size of the latent space to be $D=64$ for all models. See Appendix \ref{appn:implementation} for details. 

\noindent\textbf{Evaluation Metrics.} We use the Adjusted Rand Index (ARI)  to measure the quality of objects discovered \cite{hubert1985comparing}.
ARI is a measure of similarity between two data clusterings that takes into account the permutation-invariant nature of the predicted segmentation masks and their corresponding object ground-truth masks.
 We also use the Dice similarity coefficient, along with  the Intersection-over-Union (IoU) between the best matching object masks $X$ and $Y$:
\begin{equation}
\text{Dice}(X,Y)= \frac{2|X\cap Y|}{|X|+|Y|} \text{\hspace{0.4em};\hspace{0.4em}IoU}(X,Y)= \frac{|X\cap Y|}{|X\cup Y|} 
\end{equation}
where $X$ the set of object pixels extracted and $Y$ is the set of annotated object pixels in the ground truth. 
We compute the mean Dice and the mean IoU scores, denoted as mDice and mIoU respectively, by averaging the individual Dice and IoU scores across all matches.

\subsection{Quality of Discovered Objects}
We first evaluate the ability of ECO-Net to discover objects in images. 
Table \ref{tab:res_objdisc_sim}(a) shows the average ARI, mDice and mIoU scores based on the discovered multi-part objects in  simulated datasets.  ECO-Net achieves perfect score for all Tetrominoes test samples, and  outperforms other methods by a large margin in AbsScene with near-perfect scores.

\begin{table*}[t!]
\centering
\small
\caption{Evaluation scores of discovered foreground objects.}
		\label{tab:res_objdisc_sim}
         \vspace*{-0.1in}
        {(a) Simulated datasets}\\
%\vspace*{0.05in}
		\begin{tabular}{c|ccc|ccc}
			\hline
             &  \multicolumn{3}{c|}{Tetrominoes} &  \multicolumn{3}{c}{AbsScene} \\
            \cline{2-4}\cline{5-7}
			Method & ARI & mDice & mIoU & ARI & mDice & mIoU \\
            	\hline
            SLIC & 51.7$\pm$17.2 & 65.2$\pm$14.8 & 52.5$\pm$16.2 & 26.2$\pm$10.4 & 53.9$\pm$9.4 & 38.3$\pm$9.3 \\
            Ncut & 66.4$\pm$19.7 & 62.8$\pm$17.8 & 56.5$\pm$19.1 & 23.7$\pm$17.8 & 25.7$\pm$8.9 & 15.7$\pm$6.9 \\
            Felzenszwalb & 91.9$\pm$18.2 & 96.6$\pm$7.8 & 95.0$\pm$11.2 & 76.8$\pm$30.3 & 85.8$\pm$14.1 & 78.2$\pm$17.2 \\
            Slot Attention & 99.7$\pm$1.5 & 41.4$\pm$1.5 & 26.6$\pm$1.5 & 94.8$\pm$11.8 & 74.6$\pm$10.8 & 63.9$\pm$13.4  \\
            GENESIS-V2 & 97.2$\pm$5.4 & 47.0$\pm$3.5 & 30.9$\pm$3.0 & 46.7$\pm$24.1 & 65.4$\pm$15.8 & 54.3$\pm$18.0 \\ 
            BO-QSA & 99.1$\pm$3.7 & 40.8$\pm$1.4 & 25.8$\pm$1.2 & 93.4$\pm$13.8 & 83.1$\pm$9.2 & 74.0$\pm$12.4 \\
            DINOSAUR & 92.5$\pm$9.9 & 40.0$\pm$2.0 & 25.3$\pm$1.8 & 95.0$\pm$8.2 & 51.9$\pm$6.2 & 35.5$\pm$5.7 \\
            MaskCut & 93.4$\pm$15.3 & 80.8$\pm$5.6 & 68.5$\pm$6.6 & 90.5$\pm$18.9 & 88.3$\pm$10.7 & 81.5$\pm$12.8 \\
            OC-Net & 92.4$\pm$15.4 & 96.7$\pm$6.5 & 94.9$\pm$9.7 & 78.5$\pm$24.7 & 86.4$\pm$11.7 & 78.7$\pm$14.4 \\
			ECO-Net & \textbf{100.0$\pm$0.0} & \textbf{100.0$\pm$0.0} & \textbf{100.0$\pm$0.0} & \textbf{97.7$\pm$2.4} & \textbf{99.4$\pm$1.0} & \textbf{98.8$\pm$1.5} \\
            	\hline
		\end{tabular}

\vspace{0.2cm}
        \label{tab:res_objdisc_real}
        {(b) Realistic datasets}\\
     %   \vspace*{0.05in}
		\begin{tabular}{c|ccc|ccc}
				\hline
             &  \multicolumn{3}{c|}{GSO} &  \multicolumn{3}{c}{SKU-110K} \\
            \cline{2-4}\cline{5-7}
			Method & ARI & mDice & mIoU & ARI & mDice & mIoU \\
           	\hline
            SLIC & 18.5$\pm$16.8 & 41.1$\pm$14.0 & 27.4$\pm$12.8 & 2.7$\pm$2.0 & 1.7$\pm$0.7 & 0.8$\pm$0.3 \\
            Ncut & 29.8$\pm$2.04 & 33.4$\pm$14.0 & 23.5$\pm$12.4 & 5.2$\pm$2.6 & 8.7$\pm$4.0 & 5.5$\pm$2.8 \\
            Felzenszwalb & 61.2$\pm$34.3 & 70.3$\pm$16.9 & 58.8$\pm$18.1 & -0.8$\pm$7.4 & 7.7$\pm$2.9 & 4.9$\pm$2.0 \\
            Slot Attention & 64.3$\pm$28.8 & 39.5$\pm$13.2 & 27.0$\pm$10.8 & -9.7$\pm$7.2 & 0.9$\pm$0.4 & 0.4$\pm$0.2 \\
            GENESIS-V2 & 90.1$\pm$8.3 & 84.3$\pm$4.3 & 73.6$\pm$5.9 & 6.1$\pm$2.7 & 6.4$\pm$2.2 & 3.5$\pm$1.3 \\
            BO-QSA & 64.4$\pm$17.0 & 74.2$\pm$4.3 & 59.7$\pm$5.4 & 5.4$\pm$2.3 & 7.2$\pm$3.1 & 3.9$\pm$1.8 \\
            DINOSAUR & 96.2$\pm$10.1 & 42.0$\pm$6.2 & 26.9$\pm$5.2 & 4.4$\pm$2.3 & 12.1$\pm$3.7 & 6.8$\pm$2.3 \\
            MaskCut & 97.5$\pm$5.6 & 90.9$\pm$1.8 & 83.4$\pm$3.0 & 3.7$\pm$10.0 & 0.7$\pm$1.1 & 0.3$\pm$0.7 \\
            OC-Net & 86.6$\pm$22.9 & 86.1$\pm$10.2 & 77.9$\pm$13.1 & 7.4$\pm$3.3 & 13.2$\pm$4.3 & 7.7$\pm$2.7 \\
			ECO-Net & \textbf{99.1$\pm$1.7} & \textbf{94.5$\pm$2.1} & \textbf{89.8$\pm$3.8} & \textbf{7.8$\pm$3.2} & \textbf{31.9$\pm$9.2} & \textbf{21.6$\pm$6.8} \\
           	\hline
		\end{tabular}

\vspace{0.2cm}
        \label{tab:res_objdisc_realworld}
        {(c) Real-world datasets}\\
   %     \vspace*{0.05in}
		\begin{tabular}{c|ccc|ccc}
				\hline
             &  \multicolumn{3}{c|}{PASCAL VOC} &  \multicolumn{3}{c}{MS COCO} \\
            \cline{2-4}\cline{5-7}
			Method & ARI & mDice & mIoU & ARI & mDice & mIoU \\
           	\hline
            SLIC & 4.4$\pm$8.4 & 12.7$\pm$8.2 & 7.7$\pm$5.6 & 14.4$\pm$12.8 & 33.1$\pm$11.4 & 22.2$\pm$9.4 \\
            Ncut & 9.2$\pm$15.4 & 10.6$\pm$6.6 & 6.7$\pm$4.8 & 12.9$\pm$17.1 & 27.2$\pm$16.2 & 19.3$\pm$13.8 \\
            Felzenszwalb & 9.2$\pm$22.8 & 13.6$\pm$11.5 & 9.4$\pm$9.6 & 21.1$\pm$23.4 & 28.6$\pm$20.4 & 21.6$\pm$17.5 \\
            Slot Attention & 14.8$\pm$22.4 & 12.6$\pm$8.3 & 7.7$\pm$5.6 & 28.5$\pm$14.2 & 23.9$\pm$10.2 & 14.5$\pm$7.1 \\
            GENESIS-V2 & 13.2$\pm$20.1 & 14.6$\pm$9.5 & 9.4$\pm$6.8 & 13.2$\pm$12.5 & 21.8$\pm$9.7 & 13.9$\pm$7.2 \\
            BO-QSA & 13.0$\pm$19.5 & 13.7$\pm$8.9 & 8.6$\pm$6.2 & 24.0$\pm$15.3 & 33.3$\pm$11.6 & 22.5$\pm$9.4 \\
            DINOSAUR & 16.9$\pm$24.0 & 17.0$\pm$10.5 & 11.4$\pm$7.7 & 30.6$\pm$16.9 & 23.0$\pm$9.3 & 14.2$\pm$6.6 \\
            MaskCut & 17.7$\pm$28.5 & 19.2$\pm$15.0 & 15.5$\pm$13.7 & 37.5$\pm$23.4 & 43.0$\pm$18.6 & 33.2$\pm$16.2 \\
            OC-Net & 16.6$\pm$25.7 & 20.7$\pm$13.9 & 15.1$\pm$11.6 & 26.5$\pm$19.3 & 48.1$\pm$13.6 & 35.5$\pm$15.1 \\
			ECO-Net & \textbf{19.0$\pm$21.4} & \textbf{27.9$\pm$14.4} & \textbf{19.9$\pm$11.7} & \textbf{38.9$\pm$18.9} & \textbf{63.4$\pm$12.7} & \textbf{49.7$\pm$13.2} \\
            	\hline
		\end{tabular}
    %    \vspace{-5pt}
\end{table*}

Table \ref{tab:res_objdisc_real}(b) shows the results on realistic multi-object  datasets.
For GSO, ECO-Net shows superior performance in all metrics, illustrating its capability to effectively discover common real-world objects. 
For SKU-110K, which contains multiple small densely-packed objects, ECO-Net significantly improves the ARI scores and more than doubles the mDice and mIoU scores over all methods, demonstrating its robustness in challenging object discovery tasks.

Table \ref{tab:res_objdisc_realworld}(c) shows the results on the real-world multi-object image dataset, highlighting the scalability of our method to challenging data where objects and their parts vary significantly across instances. Improving on \cite{wu2023slotdiffusion,seitzer2022bridging}, we employ the pre-trained DINO model and 
combine the part features $\mathcal{V}$ with features obtained from DINO encoder \cite{caron2021emerging} before inputting these features into our co-part object discovery algorithm. 
ECO-Net shows superior performance for all metrics on both PASCAL VOC and MS COCO,
illustrating its ability to discover challenging, commonly seen natural objects with diverse appearances. Notably, ECO-Net outperforms all baselines by a large margin in MS COCO, demonstrating its robustness in handling a large number of object categories.

\begin{figure*}[ht!]
  \centering
  \includegraphics[width=0.94\linewidth]{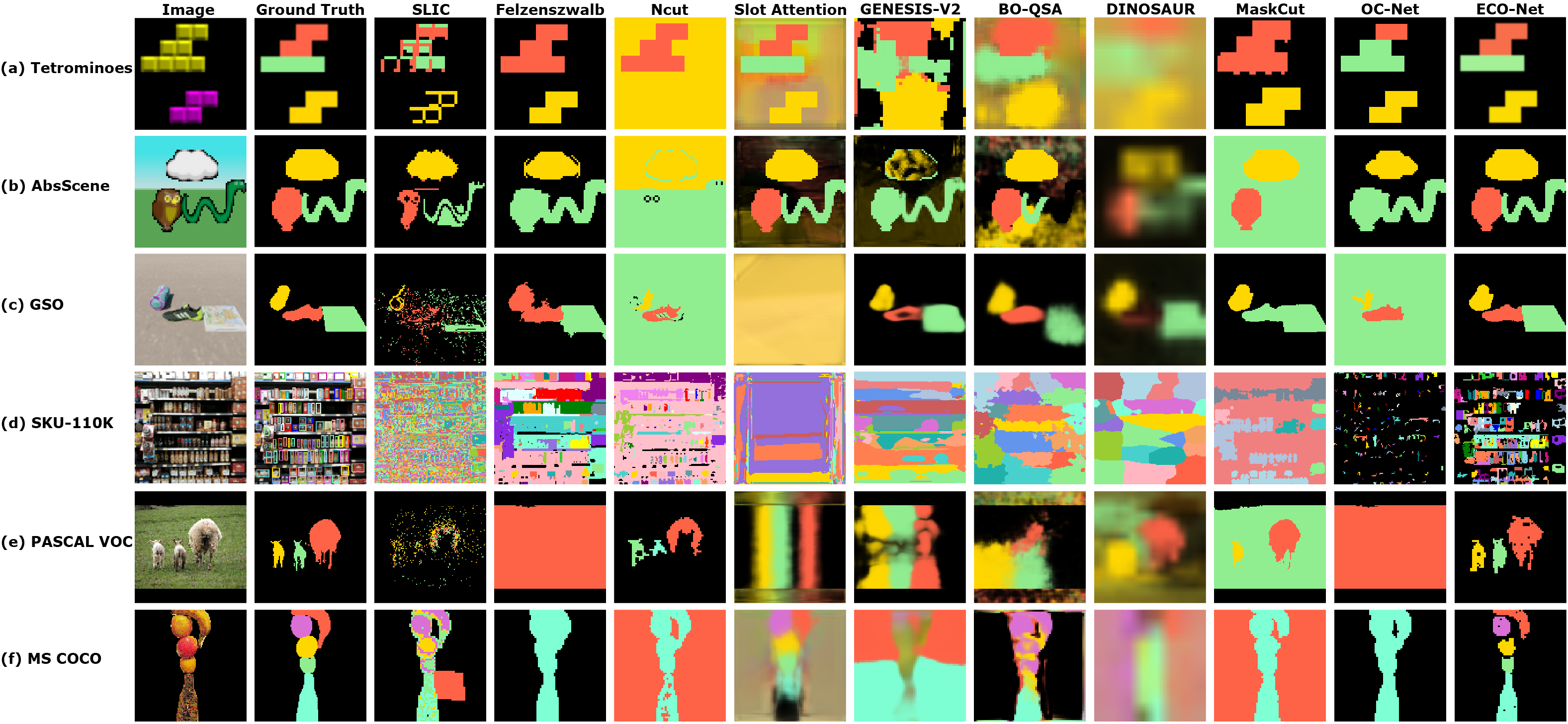}
   \caption{Visualization of discovered objects.}
   \label{fig:res_objdisc}
\end{figure*}

Figure \ref{fig:res_objdisc} visualizes the objects discovered by various methods for sample images. 
For Tetrominoes, despite the presence of scenes with connected same-color tiles that often confuse existing methods, ECO-Net still manages to separate each tile perfectly.
For AbsScene, ECO-Net manages to segment out the objects with multi-part relations in a fine-grained manner while other methods fail to consistently group together parts with varied appearances.
For GSO, only ECO-Net is able to segment out the real-world objects and their boundaries with neighboring objects in a fine-grained manner. 
For SKU-110K, ECO-Net is able to segment out the small and densely-packed objects.
For the real world datasets PASCAL VOC and MS COCO, ECO-Net is able to segment out the objects with highly varied appearances which other methods fail to discover.
Additional samples are given in Appendix \ref{appn:visuals-obj-disc}.

\begin{table*}[t!]
%\centering
\small
\begin{minipage}{0.58\textwidth}
\caption{Results for discovered objects with occlusion.}
 \vspace*{-0.1in}
	\label{tab:res_occlu}\centering
		\begin{tabular}{c|cc|cc}
				\hline
             &  \multicolumn{2}{c|}{AbsScene-O} &  \multicolumn{2}{c}{GSO-O} \\
            \cline{2-3}\cline{4-5}
			Method & mDice & mIoU & mDice & mIoU \\
            	\hline
            SLIC & 53.1$\pm$8.8 & 37.4$\pm$8.6 & 43.1$\pm$13.6 & 29.2$\pm$12.2 \\
            Ncut & 26.8$\pm$9.4 & 16.6$\pm$7.4 & 34.0$\pm$15.4 & 23.3$\pm$13.7 \\
            Felzenszwalb & 70.4$\pm$20.7 & 59.5$\pm$22.1 & 71.6$\pm$19.8 & 60.3$\pm$20.9 \\
            Slot Attention & 54.7$\pm$14.4 & 41.6$\pm$15.0 & 40.5$\pm$15.7 & 28.1$\pm$12.4  \\
            GENESIS-V2 & 48.9$\pm$19.4 & 40.3$\pm$20.6 & 81.4$\pm$5.7 & 69.4$\pm$7.0 \\ 
            BO-QSA & 61.0$\pm$13.7 & 47.1$\pm$14.9 & 72.2$\pm$6.3 & 57.3$\pm$7.6 \\
            DINOSAUR & 44.5$\pm$6.4 & 29.1$\pm$5.3 & 30.0$\pm$5.6 & 17.8$\pm$4.1 \\
            MaskCut & 68.6$\pm$21.8 & 59.1$\pm$23.9 & 85.8$\pm$3.6 & 76.9$\pm$4.6 \\
            OC-Net & 69.9$\pm$21.0 & 59.2$\pm$22.5 & 84.5$\pm$5.3 & 76.2$\pm$7.0 \\
			ECO-Net & \textbf{93.1$\pm$9.6} & \textbf{90.5$\pm$12.7} & \textbf{90.3$\pm$6.7} & \textbf{83.3$\pm$9.1} \\
            	\hline
		\end{tabular}
        % \end{table}
\end{minipage}
\begin{minipage}{0.4\textwidth}
%\centering
% \begin{table}[ht!]
\small
\caption{Results for discovered objects on out-of-distribution test set AbsScene-C.}
 \vspace*{-0.1in}
		\label{tab:res_gen} \centering
		\begin{tabular}{c|ccc}
				\hline
			Method & ARI & mDice & mIoU \\
            	\hline
            SLIC & 21.0$\pm$8.6 & 45.8$\pm$8.9 & 30.8$\pm$8.0 \\
            Ncut & 40.6$\pm$23.6 & 51.7$\pm$18.9 & 40.9$\pm$19.4 \\
            Felzenszwalb & 60.7$\pm$34.0 & 68.6$\pm$25.1 & 60.8$\pm$27.1 \\
            Slot Attention & 37.9$\pm$21.0 & 41.4$\pm$12.4 & 28.0$\pm$11.5 \\
            GENESIS-V2 & 23.2$\pm$14.3 & 32.9$\pm$8.3 & 20.5$\pm$6.3 \\ 
            BO-QSA & 32.5$\pm$16.9 & 38.3$\pm$10.0 & 25.0$\pm$8.6 \\
            DINOSAUR & 82.0$\pm$17.5 & 54.2$\pm$6.3 & 37.8$\pm$6.0 \\
            MaskCut & 67.0$\pm$39.9 & 74.9$\pm$17.8 & 64.0$\pm$20.0 \\
            OC-Net & 60.8$\pm$34.2 & 69.0$\pm$24.5 & 61.1$\pm$26.7 \\
			ECO-Net & \textbf{93.5$\pm$7.5} & \textbf{97.3$\pm$4.1} & \textbf{95.4$\pm$6.5} \\
            	\hline
		\end{tabular}
\end{minipage}
\end{table*}

\begin{table*}[h!]
\small
    \caption{$R^2$ scores for object property prediction on simulated datasets.}
    \vspace*{-0.1in}
	\begin{center}
		\label{tab:res_objrep}
		\begin{tabular}{c|cccccc|ccc}
				\hline
            & \multicolumn{6}{c}{Tetrominoes} & \multicolumn{3}{c}{AbsScene}\\
            \cline{2-7}\cline{8-10}
			Method & Red & Green & Blue & X-coord & Y-coord & Shape & X-coord & Y-coord & Shape\\
            	\hline
            Slot Attention & 85.0 & 80.8 & 93.5 & 99.1 & 98.3 & 36.2 & 97.2 & 96.3 & 76.1 \\
            GENESIS-V2 & 84.5 & 94.7 & 84.7 & 96.3 & 92.7 & 37.9 & 95.5 & 38.3 & 77.5 \\
            BO-QSA & 97.3 & 98.4 & 98.6 & 98.7 & 99.0 & 52.4 & 90.1 & 98.1 & 96.2 \\
            DINOSAUR & 91.2 & 89.6 & 86.9 & 93.1 & 94.3 & 20.1 & 97.5 & 98.0 & 95.4 \\
            OC-Net & 91.2 & 92.3 & 91.8 & 89.7 & 89.9 & 81.8 & 90.7 & 90.4 & 95.7\\
			ECO-Net & \textbf{100.0} & \textbf{100.0} & \textbf{100.0} & \textbf{99.5} & \textbf{99.4} & \textbf{98.7} & \textbf{98.5} & \textbf{98.1} & \textbf{99.7} \\
           	\hline
		\end{tabular}
	\end{center}
\end{table*}

\begin{table*}[t!]
\small
\centering
\caption{mIoU scores for variants of ECO-Net.}
    \label{tab:res_objdisc_abs}
    \begin{tabular}{ccccccccccc}
        \toprule
         Method & \multicolumn{1}{c}{Tetrominoes} & \multicolumn{1}{c}{AbsScene}  & \multicolumn{1}{c}{GSO} & \multicolumn{1}{c}{SKU-110K} & \multicolumn{1}{c}{PASCAL} & \multicolumn{1}{c}{COCO} &
         \multicolumn{1}{c}{AbsScene-O} &
         \multicolumn{1}{c}{GSO-O} &
         \multicolumn{1}{c}{AbsScene-C} \\
        \midrule
        ECO-Net & \textbf{100.0$\pm$0.0} & \textbf{98.8$\pm$1.5} & \textbf{89.8$\pm$3.8} & \textbf{21.6$\pm$6.8} & \textbf{19.9$\pm$11.7} & \textbf{49.7$\pm$13.2} & 
        \textbf{90.5$\pm$12.7} & \textbf{83.3$\pm$9.1} &
        \textbf{95.4$\pm$6.5}\\
        w/ pretrained & 57.3$\pm$12.7 & 72.8$\pm$16.7 & 83.5$\pm$3.0
        & 2.8$\pm$0.4 & \textbf{19.9$\pm$11.7} & \textbf{49.7$\pm$13.2}
        & 58.0$\pm$21.6 & 79.8$\pm$12.6 & 60.1$\pm$20.3\\
        w/o  co-part
        & 40.6$\pm$29.2 & 61.6$\pm$23.4 & 14.7$\pm$13.9
        & 7.8$\pm$2.8 & 5.9$\pm$6.3 & 6.8$\pm$9.3
        & 51.6$\pm$20.1 & 25.1$\pm$25.8 & 31.9$\pm$27.1\\
        \bottomrule
    \end{tabular}
\end{table*}

\subsection{Experiments on Occlusion-Aware Perception}
Occlusion-aware perception evaluates how effectively an object-centric model can recognize objects when they are only partially visible.
Importantly, when a model encounters an object with partially visible appearance, it should  be able to identify the object based on memory. This can be quantified by measuring how well the model `fills-in' the occluded parts. We assess the occlusion-aware perception of ECO-Net with the baselines by training  on AbsScene and GSO, and testing on AbsScene-O and GSO-O respectively, using  ground-truth objects with occluded parts completed.

\begin{figure*}[h!]
  \centering \small
  {\includegraphics[width=0.94\linewidth]{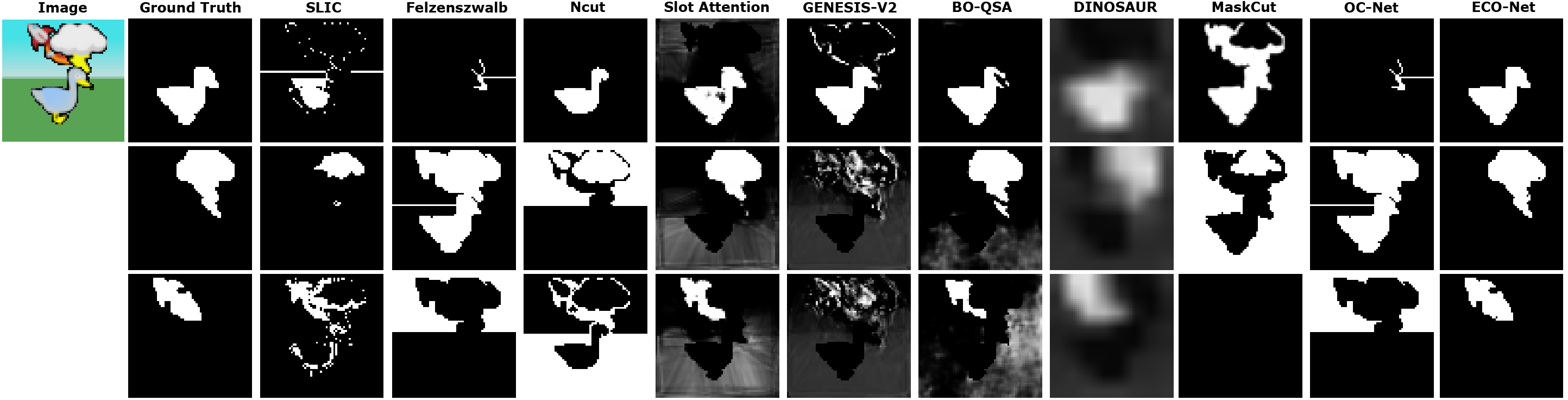} \label{fig:res_occlu_sim_absscene}}\\ 
  (a) AbsScene-O\\
 {\includegraphics[width=0.94\linewidth]{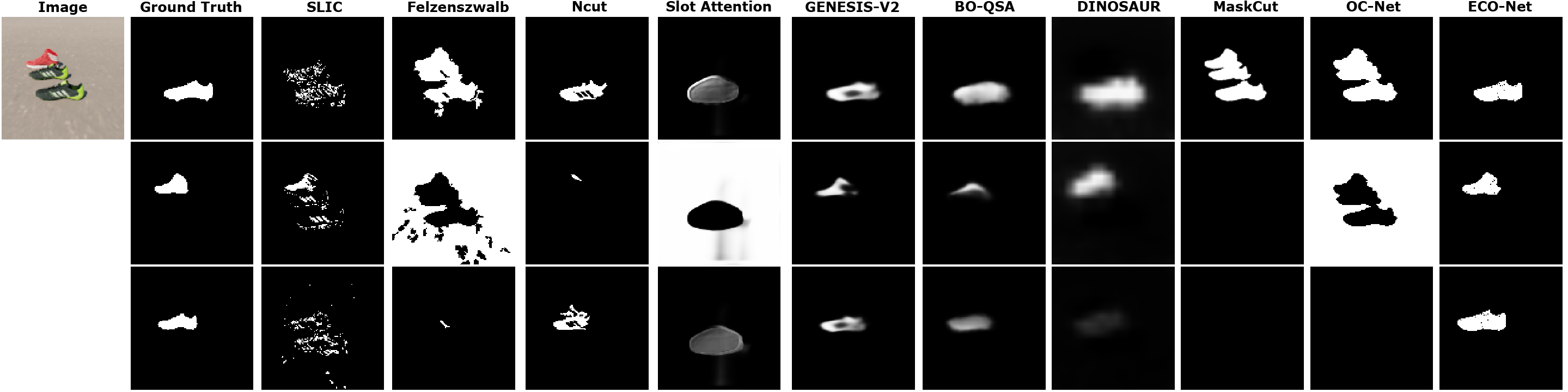} \label{fig:res_occlu_real_gso}}\\
  (b) GSO-O
  \caption{Visualization of discovered objects with occlusion.} \label{fig:res_occlu} 
\end{figure*}

Table \ref{tab:res_occlu} shows the results with ECO-Net outperforming all baselines by a large margin. Figure \ref{fig:res_occlu} and Appendix \ref{appn:visuals-occlu} shows the occluded objects discovered by the various methods. We see that the overlapping of multi-part objects continue to confuse existing SOTA. Notably, BO-QSA and DINOSAUR struggle to discern the borders of overlapping objects while MaskCut and OC-Net fail to separate the objects for both AbsScene-O and GSO-O samples. In contrast, ECO-Net discovers complete objects accurately.

\subsection{Out-of-Distribution Discovery}

An effective object-centric representation should generalize across different scenes by accurately encoding object appearances. 
Here, we demonstrate that the object representations learned from AbsScene by ECO-Net are robust and can be used to discover objects in AbsScene-C, which consists of 
 random complex-texture backgrounds.

\begin{figure*}[h!]
  \centering
  {\includegraphics[width=0.99\linewidth]{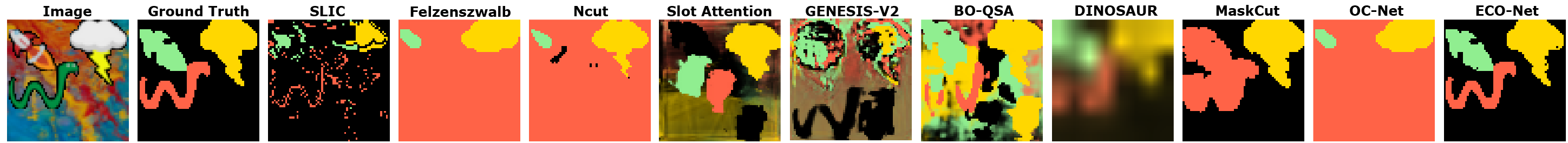}}
  \caption{Visualization of discovered objects on out-of-distribution test set AbsScene-C.} \label{fig:res_gen} 
\end{figure*}

Table \ref{tab:res_gen} shows a general performance decline across  all models due to the shift in data distribution. However, 
 ECO-Net has the smallest drop in performance and significantly surpasses the best performing method by a large margin. 
Figure \ref{fig:res_gen} illustrates how, despite the complex-textured backgrounds that confuse other state-of-the-art methods, ECO-Net remains robust in discovering the complete objects. Additional visualizations are included in Appendix \ref{appn:visuals-ood}.

\subsection{Experiments on Object Property Prediction}
A key trait of effective object-centric representation is its ability to encode object properties such as color, position and shape \cite{Scholkopfetal21}. 
Here, we show that  the learned object representations are disentangled and can be used to predict the values of these properties. 

Given the representations and their corresponding ground truth values of a target object property, we employ them as features to train a gradient boosted tree (GBT).
To evaluate how well the properties of unseen objects are  predicted by the GBT, we use the coefficient of determination $R^2$ \cite{Wright1921CorrelationAndCausation}.
We use the learned object representations from the simulated datasets to predict the properties of objects. We use the mIoU score to match the discovered object to the ground truth object. We split the test images equally into two sets, one for training the GBT model for each object property, and the other for evaluation.

Table \ref{tab:res_objrep} shows the average $R^2$ scores of GBT models on the evaluation set. 
GBT models trained with ECO-Net representations achieved the highest $R^2$ scores compared to models trained using the representations from baselines. This suggests that the object representations learned by ECO-Net are effective in encoding object properties.

\subsection{Ablation Studies}
Next, we examine the effect of co-part object discovery and the use of pretrained features on the performance of ECO-Net. 
We implemented two variants of ECO-Net: 
 (a) w/ pretrained. This network combines the part features $\mathcal{V}$ with features obtained from pretrained DINO encoder before inputting these features into the co-part object discovery algorithm.
 (b) w/o co-part. Here, we do not perform the clustering of parts into object wholes.

Table \ref{tab:res_objdisc_abs} shows the mIoU scores for all the datasets.
With pretrained features, we observe a drop in performance across almost all datasets since a reliance on the pretraining distribution causes an inconsistent discovery of parts from the same object. As such, ECO-Net w/ pretrained struggles to cluster parts accurately into object wholes. 

Removing the co-part object discovery module (ECO-Net w/o co-part) leads to a large performance drop in all datasets, indicating the importance of leveraging on parts to consistently and accurately discover multi-part objects especially in occluded and out-of-distribution settings.

\section{Conclusion}
We have described a framework that leverages on explicit graph representations for parts and introduce a co-part object discovery algorithm to iteratively build up parts into objects.
Experiments on simulated, realistic and real-world datasets have demonstrated the superior quality of the objects discovered over state-of-the-art. 
We have also highlighted that our solution excels in occlusion-aware perception and out-of-distribution generalization. 
Finally, we show that the obtained object representations can predict key properties in a downstream task, indicating its potential use for applications where samples and labels are limited.

{
    \small
    \bibliographystyle{ieeenat_fullname}
    \bibliography{main}
}

\clearpage
\newpage
\appendix

% \onecolumn
\section{Tractable Co-Part Object Discovery}
\label{appn:copart}
In this section, we provide the proof to Theorem \ref{theo:consubgraph} by formally analyzing the completeness and tractability of co-part object discovery via pairwise comparisons of neighboring parts.

\begin{proof}
Let $\hat{\mathcal{G}} = (\mathcal{P}, \hat{\mathcal{E}})$ be a graph of parts from an input batch of images, where $\mathcal{P} = \{P_1, \dots, P_M\}$ is the set of parts with features $\mathcal{V} = \{\mathbf{V}_1, \dots, \mathbf{V}_M\}$, and $\hat{\mathcal{E}} = \{(P_i, P_j)\}$, where $1 \leq i < j \leq M$, is the set of edges with corresponding spatial labels. Define the graph of adjacent parts $\mathcal{G} = (\mathcal{P}, \mathcal{E})$, where $\mathcal{E}\subset\hat{\mathcal{E}}$, and parts are adjacent if they share a boundary of nonzero length (i.e., there exists at least one pair of adjacent pixels).

Consider the following pairwise iterative algorithm that clusters neighboring parts with similar relations into objects:

\begin{enumerate}
    \item \textbf{Pairwise Comparison:} For each pair of parts $[P_i, P_j]$, where $1 \leq i < j \leq M$, perform the following checks:
    \begin{itemize}
        \item Compare the features $\mathbf{V}_i$ and $\mathbf{V}_j$ for similarity.
        \item Compare the similarity of their edges with neighboring parts.
        \item For each similar edge $(P_i, P_k) \in \mathcal{E}$ and $(P_j, P_{k'}) \in \mathcal{E}$, compare the features of the incident parts $P_k$ and $P_{k'}$.
    \end{itemize}
    
    \item \textbf{Object Assignment:} If all three similarity conditions are satisfied for a pair $[P_i, P_j]$ and their incident parts $[P_k, P_{k'}]$, proceed as follows:
    \begin{enumerate}
        \item If an object already contains the pair $[P_i, P_j]$, add the incident parts $[P_k, P_{k'}]$ and their respective edges to the object. Complete edges with other adjacent parts and merge other objects containing the pair.
        \item If an object contains the pair of incident parts $[P_k, P_{k'}]$, add pair $[P_i, P_j]$ and their respective edges to the object. Complete edges with other parts and merge other objects containing the pair.
        \item If no object contains either the pair $[P_i, P_j]$ or the incident parts $[P_k, P_{k'}]$, create a new object containing these parts and edges.
    \end{enumerate}
    
    \item \textbf{Iteration:} Repeat the above steps for every pairs of parts until all pairs have been considered.
\end{enumerate}

\textbf{Proof of Completeness:}
Since the algorithm systematically examines every possible pair of parts $[P_i, P_j]$ and, for each, considers all possible incident part pairs $[P_k, P_{k'}]$, it exhaustively enumerates all potential similarity matches between parts and their relations. 
For each configuration, the algorithm checks all relevant constraints (features, edges, incident parts, merge conditions) and updates the object assignments accordingly. This ensures that:
\begin{itemize}
    \item No possible matching of similar parts and their incident relations is omitted,
    \item All constraints and complete adjacencies are verified for each candidate grouping,
    \item Objects are incrementally and correctly built up from all valid pairwise and incident-part relations.
\end{itemize}

Therefore, the algorithm guarantees completeness in clustering: every valid grouping of parts into objects, according to the specified similarity and relational constraints, will be found.

\bigskip

\textbf{Proof of Tractability:}
Let $|\mathcal{P}| = M$ be the number of parts (segments), and let $d$ be the maximum degree (number of neighbors) of a part in the graph of parts $\mathcal{G}$.

\begin{enumerate}
    \item \textbf{Pairwise Comparison.}
    \begin{itemize}
        \item The number of unordered pairs $[P_i, P_j]$ is $\binom{M}{2} = O(M^2)$.
        \item For each pair:
        \begin{itemize}
            \item Comparing features $\mathbf{V}_i$ and $\mathbf{V}_j$ is $O(1)$.
            \item Comparing the similarity of their neighbor edges involves up to $d$ neighbors each, i.e., $O(d^2)$ operations per pair.
            \item For each such neighbor pair $((P_i, P_k), (P_j, P_{k'}))$, comparing $P_k$ and $P_{k'}$ is $O(1)$.
        \end{itemize}
        \item Total work per pair is $O(d^2)$.
    \end{itemize}
    
    \item \textbf{Object Assignment.}
    \begin{itemize}
        \item Per pair, object lookup, insertion, and merging (with efficient data structures like hash tables and union-find) are $O(\log M)$ amortized.
        \item Completing edges with other adjacent parts in the object is $O(d)$ per update.
        \item Across all pairs, the total cost for this step is $O(M^2 \log M)$ at worst.
    \end{itemize}
    
    \item \textbf{Total Time Complexity:}
    $
    O(M^2 d^2) + O(M^2 \log M)
    $
\end{enumerate}

We now consider the theoretical bound on the maximum degree $d$ of the input graph of adjacent parts $\mathcal{G} = (\mathcal{P}, \mathcal{E})$.

\paragraph{Planarity of Graph-Based Segmentation:}
Let the segmented parts $\mathcal{P}$ be obtained via a graph-based image segmentation algorithm such as Felzenszwalb’s algorithm~\cite{felzenszwalb2004efficient}. 
In this approach, the input image is represented as a graph where each pixel is a vertex and edges connect spatially neighboring pixels; this initial pixel grid graph is planar by construction. 
The segmentation algorithm merges pixels into segments according to a boundary predicate (see Section~3.1 in~\cite{felzenszwalb2004efficient}), yielding at every stage contiguous segments, each corresponding to a connected region in the plane.

After segmentation, we define adjacency between parts such that two parts are adjacent if they share a boundary of nonzero length. The resulting graph of adjacent parts, $\mathcal{G}$, is thus equivalent to the dual graph of the planar subdivision defined by the segmentation. By classical graph theory, the dual graph of any partition of the plane into regions is itself planar~\cite{bondy2008graph}.

For any planar graph with $M$ vertices, Euler’s formula and its corollaries guarantee that the average degree is strictly less than $6$ \cite{bondy2008graph}; thus, $d = O(1)$ except in highly contrived constructions. In practice, for natural images and reasonable segmentation parameters, the observed $d$ is much smaller (typically $4$--$8$).

\vspace{1em}
\textbf{Summary of Cases:}
\begin{itemize}
    \item \textbf{Theoretical guarantee:} By planarity, the average degree is strictly less than $6$, so $d = O(1)$ except in pathological cases.
    \item \textbf{Worst-case theory:} The algorithm does not impose a strict upper bound on $d$; in artificially constructed segmentations, $d$ could be as large as $M-1$, resulting in $O(M^4)$ complexity.
\end{itemize}

\textbf{Conclusion:} \\
Because the graph of adjacent parts $\mathcal{G}=(\mathcal{P},\mathcal{E})$ is planar and planar graph theory ensures a small degree \cite{bondy2008graph}, and because Felzenszwalb’s algorithm empirically produces compact, contiguous segments, the considered algorithm is tractable in both theory and practice for planar image segmentations:
$
O(M^2 d^2) + O(M^2 \log M) = O(M^2 \log M) = O(|\mathcal{P}|^2 \log |\mathcal{P}|)
$
for standard images and segmentations, with only rare, artificially constructed cases approaching $O(M^4)$.

\end{proof}

\newpage
\section{Co-part Object Discovery Algorithm}
\label{co-part-objdisc}
We design a co-part object discovery algorithm which iteratively clusters recurrent parts into objects. Details of the algorithm is given below. 
\begin{algorithm}[H]
   \small
   \label{co-part-objdisc-code}
   \caption{Co-Part Object Discovery.}
\begin{algorithmic}[1]
   \STATE {\bfseries Input:} Set of parts $\mathcal{P} =\{P_1,\dots,P_M\}$; set of part features $\mathcal{V} =\{\mathbf{V}_1,\dots,\mathbf{V}_M\}$; set of edges for adjacent parts $\mathcal{E} = \{(P_1,P_1),\dots,(P_M,P_M)\}$
   \STATE {\bfseries Output:} Set of objects $\mathcal{O} = \{\mathcal{O}_1,\dots,\mathcal{O}_c\}$\\
   \STATE $\mathcal{U} \gets \mathcal{P}$; $\mathcal{O} \gets \emptyset$;
    $c$ $\gets$ $1$ \hspace{1em} // Initialization 
   \REPEAT
   \STATE $P_i \gets$ UniformSampling($\mathcal{U}$)
   \STATE let $\mathcal{V}' \subset \mathcal{V} - \{\mathbf{V}_i\}$ be the set of part features with similarity more than $\mathbf{\epsilon}$ to $\mathbf{V}_i$
   \FOR{each $\mathbf{V}_j \in \mathcal{V}'$}
   \STATE let $\mathcal{E}_i,\mathcal{E}_j \subset \mathcal{E}$ be the set of edges connected to parts $P_i$ and $P_j$ respectively
   % \FOR{each edge $(P_j,P_k)$ connected to $\mathcal{P}_j$}
   \FOR{each pair of edges $((P_i,P_{k}),(P_j,P_{k'})) \in \mathcal{E}_{i,j} \subset \mathcal{E}\times\mathcal{E}$ with similarity more than $\epsilon$}
   % \IF {sim$((P_j,P_k), (P_{i},P_{k'})) > \epsilon$}
   \IF {sim$(\mathbf{V}_k,\mathbf{V}_{k'}) > \epsilon$}
   \IF {$\mathbf{V}_i$,$\mathbf{V}_k$ do not exist in some object}
   \STATE $\mathcal{P}_c \gets \{\mathbf{V}_i,\mathbf{V}_k\}$
   \STATE $\mathcal{E}_c \gets \{(P_i,P_k)\}$
   \STATE $\mathcal{O}_c \gets \{\mathcal{P}_c,\mathcal{E}_c\}$
   \STATE $\mathcal{O} \gets \mathcal{O} \cup \{\mathcal{O}_c\}$; $c$ $\gets$ $c + 1$
   \ELSE
   \STATE let $\mathcal{O}_c = \{\mathcal{P}_c,\mathcal{E}_c\}$ be the object that contains $\mathbf{V}_i$ or $\mathbf{V}_k$
   \STATE $\mathcal{P}_c \gets \mathcal{P}_c \cup \{\mathbf{V}_i,\mathbf{V}_k\}$
   \STATE $\mathcal{E}_c \gets \mathcal{E}_c \cup \{(P_i,P_k)\}$
   \STATE For each $P_{y} \in \mathcal{P}_c$, if $(P_y, P_{i}) \in \mathcal{E}$, add to $\mathcal{E}_c$; same for $P_{k}$
   \STATE For any other objects $\mathcal{O}_y$ containing $\mathbf{V}_i$ or $\mathbf{V}_k$ respectively, merge all such into $\mathcal{O}_c$
   \ENDIF
   \ENDIF
   \ENDFOR
   \ENDFOR
   \STATE $\mathcal{U} \gets \mathcal{U} - \{P_i\}$
   \UNTIL{$|\mathcal{U}| == \emptyset$}
\end{algorithmic}
\end{algorithm}

\newpage
\section{Datasets}
\label{appn:datasets}
\begin{enumerate}
    \item Tetrominoes \cite{multiobjectdatasets19} (Apache-2.0 License): This dataset has $35\times 35$ images, each consisting of 3 Tetris-like shapes sampled from 6 colors and 17 shapes. We download the data from the original github page\footnote{\textit{https://github.com/deepmind/multi\_object\_datasets}}. Ground truth segmentation masks and properties are provided.
    \item Abstract Scenes (AbsScene) \cite{zitnick2013bringing}: This dataset consists of $64\times 64$ images with 10 multi-part objects. Each object consists of multiple parts with different appearances, and objects vary in position across images. We download the data from the original release\footnote{\textit{https://www.microsoft.com/en-sg/download/details.aspx?id=52035}}. Ground truth segmentation masks are provided.
    \item Abstract Scenes with Occlusion (AbsScene-O) \cite{zitnick2013bringing}: We generated this variation of the AbsScene dataset, where occlusion of objects is allowed up to $25\%$ minimum visibility for each object. Ground truth masks of the full objects are provided.
    \item Abstract Scenes with Complex Backgrounds (AbsScene-C) \cite{zitnick2013bringing}: We generated this out-of-distribution test set of the AbsScene dataset, where we randomly replace the original background with 15 real-world complex-textured backgrounds. Ground truth masks of the full objects are provided.
    \item Google Scanned Objects (GSO) \cite{downs2022google}. We generated this dataset which consists of photo-realistic images made via compositions of common real-world objects and backgrounds. Each generated image contains realistic variations of lighting, shadows, 3D object positions and object-occlusion. All raw images and their ground truths are resized and cropped to size $128\times 128$. We downloaded the raw data from the official website\footnote{\textit{https://app.gazebosim.org/GoogleResearch/fuel/collections/\\Scanned\%20Objects\%20by\%20Google\%20Research}}, and generated the dataset by using the Kubric setup following the official github repository\footnote{\textit{https://github.com/jinyangyuan/compositional-scene-representation-datasets}} from \cite{yuan2022compositional}.
    \item  Google Scanned Objects with Occlusion (GSO-O) \cite{downs2022google}: We generated this variation of the GSO dataset, where occlusion of objects is allowed up to $25\%$ minimum visibility for each object. Ground truth masks of the full objects are provided.
    \item  SKU-110K (SKU) \cite{goldman2019precise}: This challenging real-world dataset provides 11,762 images with 1.7 million annotated objects captured in densely packed store-keeping scenarios. The images are collected from supermarket stores and are of various scales, viewing angles, lighting conditions, and noise levels. All the images and their ground truth bounding boxes are cropped and resized into size $128\times 128$. We download the data from the official github page\footnote{\textit{https://github.com/eg4000/SKU110K\_CVPR19}}.
    \item PASCAL VOC \cite{pascal-voc-2012}: The PASCAL VOC (Visual Object Classes) challenge 2012 dataset is a widely used benchmark for object segmentation with ground truth annotations for 20 object classes and 1 background class. The images are collected from real-world settings and contain objects which are in natural environments of various scales, viewing angles, lighting conditions, and noise levels. We train on the trainaug variant with 10582 images and we evaluate on the trainaug validation set with 1449 images. All the images and their ground truth bounding boxes are cropped and resized into size $128\times 128$. We download the data from the official website\footnote{\textit{http://host.robots.ox.ac.uk/pascal/VOC/voc2012/index.html}}.
    \item MS COCO  \cite{lin2014microsoft}: The MS COCO (Microsoft Common Objects in Context) dataset is a large-scale dataset for object segmentation, containing over 330,000 images annotated with 80 object categories. We use the variant that focuses on multi-object segmentation as proposed by \cite{yang2022promising}, which we download from here\footnote{\textit{https://www.dropbox.com/sh/u1p1d6hysjxqauy/AACgEh0K5A\\NipuIeDnmaC5mQa?dl=0}}. All the images and their ground truth bounding boxes are cropped and resized into size $128\times 128$.
\end{enumerate}

\section{Baseline Models}
\label{appn:baselines}
\begin{enumerate}
    \item SLIC \cite{SLIC} is a clustering algorithm that clusters pixels into superpixels by using an efficient adaptation of the k-means algorithm. We use the python implementation and for each dataset, we perform a grid search for the optimal hyperparameters which produce the best results. For example, for the Tetrominoes dataset, we set the optimal clustering threshold value as $10$ and the initial number of clusters as $12$.
    \item Felzenszwalb's Algorithm \cite{emami2021efficient} is a graph-based segmentation algorithm that groups pixels together through a hand-crafted boundary detection procedure. We use the python implementation and for each dataset, we perform a grid search for the optimal hyperparameters which produce the best results. For example, for the Tetrominoes dataset, we set the optimal clustering threshold $1000$, the minimum cluster size as $10$, and the image smoothening value as $0.1$.
    \item Normalized Cut (Ncut) algorithm~\cite{shi2000normalized} is an unsupervised image segmentation algorithm which treats image segmentation as a graph partitioning problem and uses a hand-crafted criterion to measure both the total dissimilarity between the different groups as well as the total similarity within the groups in order to determine the final segmentation groups.
    \item Slot Attention \cite{locatello2020object} initializes a set of random object representations called slots which are iteratively refined by slot-normalized cross-attention on the outputs of a simple convolutional neural network (CNN). The slots are then decoded individually and combined to reconstruct the input. All Slot Attention baselines are trained with 300,000 iterations. We use the default training hyperparameters from the official reference implementation.
    \item GENESIS-V2 \cite{engelcke2021genesis} obtains pixel embeddings through a U-Net which are then clustered using a stochastic stick-breaking process. The clusters are then decoded to reconstruct the input. We similarly fine-tune the model to use the output standard deviation of 0.7 and the equivalent per-pixel GECO reconstruction target as EfficientMORL and achieved much higher results than those reported in the CLEVRTEX paper \cite{clevrtex}.
    \item BO-QSA \cite{jia2023improving} initializes Slot Attention’s object representations as learnable embeddings instead of sampling from a learnable Gaussian distribution and supplements the training with bi-level optimization. All BO-QSA baselines are trained with 300,000 iterations. We use the default training hyperparameters from the official reference implementation. We train all datasets with both the mixture-based decoder and autoregressive transformer-based decoder and report the highest scores. 
    \item DINOSAUR \cite{seitzer2022bridging} leverages on pre-trained features and Slot Attention to scale to real-world images. DINOSAUR first extracts learned features of the input image by using the DINO model, which was pre-trained on the ImageNet dataset using self-supervised learning techniques \cite{caron2021emerging}, then uses Slot Attention to attend to different sets of features before reconstructing the self-supervised features. All DINOSAUR baselines are trained with 300,000 iterations. We use the default training hyperparameters from the validated reference implementation.
    \item MaskCut~\cite{wang2023cut} was recently proposed to perform unsupervised image segmentation by leveraging on a pre-trained model. MaskCut first extracts learned features of the input image by using the DINO model, which was pre-trained on the ImageNet dataset using self-supervised learning techniques \cite{caron2021emerging}, before applying the Ncut algorithm on the extracted features to determine the segmentation. 
    \item OC-Net \cite{foo2023multi} leverages on learned feature connectivity to discover objects in an unsupervised manner. OC-Net also uses two designed object-centric regularization terms to ensure good separation between object representations and suitable disentanglement within representations. We use similar training iterations, thresholds and loss weights as proposed in the original paper for all the datasets.
\end{enumerate}

\section{Additional Implementation Details}
\label{appn:implementation}

\subsection{Hyperparameters}
Our ECO-Net framework has 2 main hyperparameters: latent-space size $K$ and similarity constant $\epsilon$. For latent-space size $K$ which determines the number of sampled vectors from each part centroid, we follow the baseline models and set the representation latent-space size to $K=64$. For the similarity constant $\epsilon$, we describe our initial experiment to determine its value as follows: we first generate 2 samples of the AbsScene dataset, similar to the input images in Figure \ref{fig:method_overview}. After running the co-part object discovery algorithm with grid searched values of $\epsilon$, we find that the object can be accurately discovered when $0.2 \leq \epsilon \leq 1$. Considering that a lower $\epsilon$ increases the number of similarity matches which increases the runtime, we set $\epsilon=0.99$ for efficiency and accuracy.

\subsection{Object Similarity Computations}
Our ECO-Net framework uses a graph-based representation for objects, where each object in memory is defined by its part nodes, part edges and object node. Nodes are derived by sampling $K$ vectors which point from the respective centroid, while edges are vectors which point from the centroid of one part to an adjacent part. In order to determine whether a part or an object is similar to another, we compute the similarity (inverse distance) between the part nodes, part edges and object node. We then take the mean. We have attached a simple demo for this computation in the supplementary material.

\subsection{Settings for Felzenszwalb's Algorithm}
We use Felzenszwalb's algorithm to obtain the parts in ECO-Net's Graph Representation Module. 
Since we are concerned with fine-grained parts formed by clustering together pixels with similar features, we fix the threshold to $10$ and the minimum cluster size as $1$ for all experiments.

\subsection{Third-Party Assets}
ECO-Net is implemented using PyTorch \cite{NEURIPS2019_9015}. In addition to various open-sourced Python packages, we make use of the following third-party assets:
\begin{itemize}
    \item \citet{locatello2020object} (Apache-2.0 license): Implementation of Slot Attention\footnote{\textit{https://github.com/google-research/google-research/tree/master/slot\_attention}},
    \item \citet{engelcke2021genesis} (GPL-3.0 license): Implementation of GENESIS-V2\footnote{\textit{https://github.com/applied-ai-lab/genesis}},
    \item \citet{jia2023improving}: Implementation of BO-QSA\footnote{\textit{https://github.com/YuLiu-LY/BO-QSA}},
    \item \citet{seitzer2022bridging} (MIT License): Implementation of DINOSAUR\footnote{\textit{https://github.com/gorkaydemir/DINOSAUR}}.
    \item \citet{wang2023cut} (CC-BY-NC 4.0 License): Implementation of MaskCut\footnote{\textit{https://github.com/facebookresearch/CutLER}}.
\end{itemize}

\subsection{Computational Resources}
\label{compute}
An 8X Tesla V100 (32GB) GPU server is used to train ECO-Net and all comparison baselines for our experiments.

\section{Additional Visualizations}
\subsection{Co-Part Object Discovery Algorithm}
\begin{figure*}[h!]
  \centering
  \subfloat[Sample 1]{\includegraphics[width=0.85\linewidth]{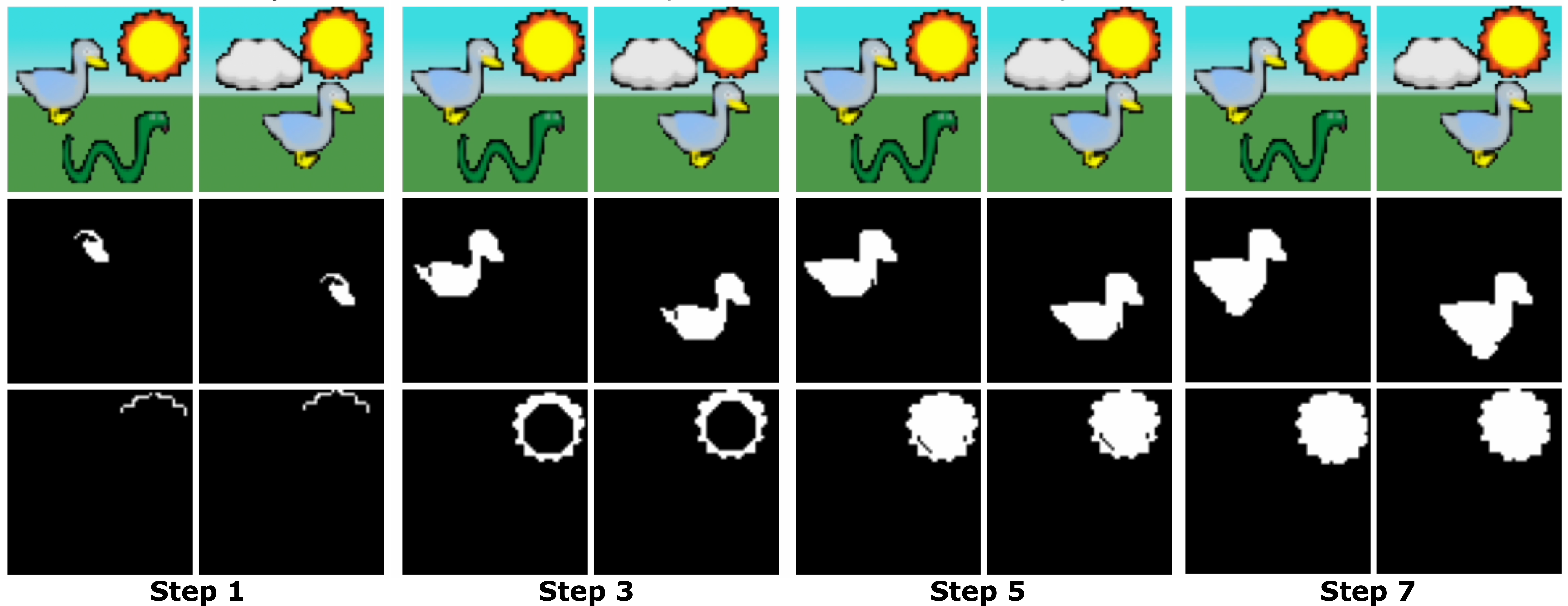} \label{fig:res_objdisc_demo_0}}\\
  \subfloat[Sample 2]{\includegraphics[width=0.85\linewidth]{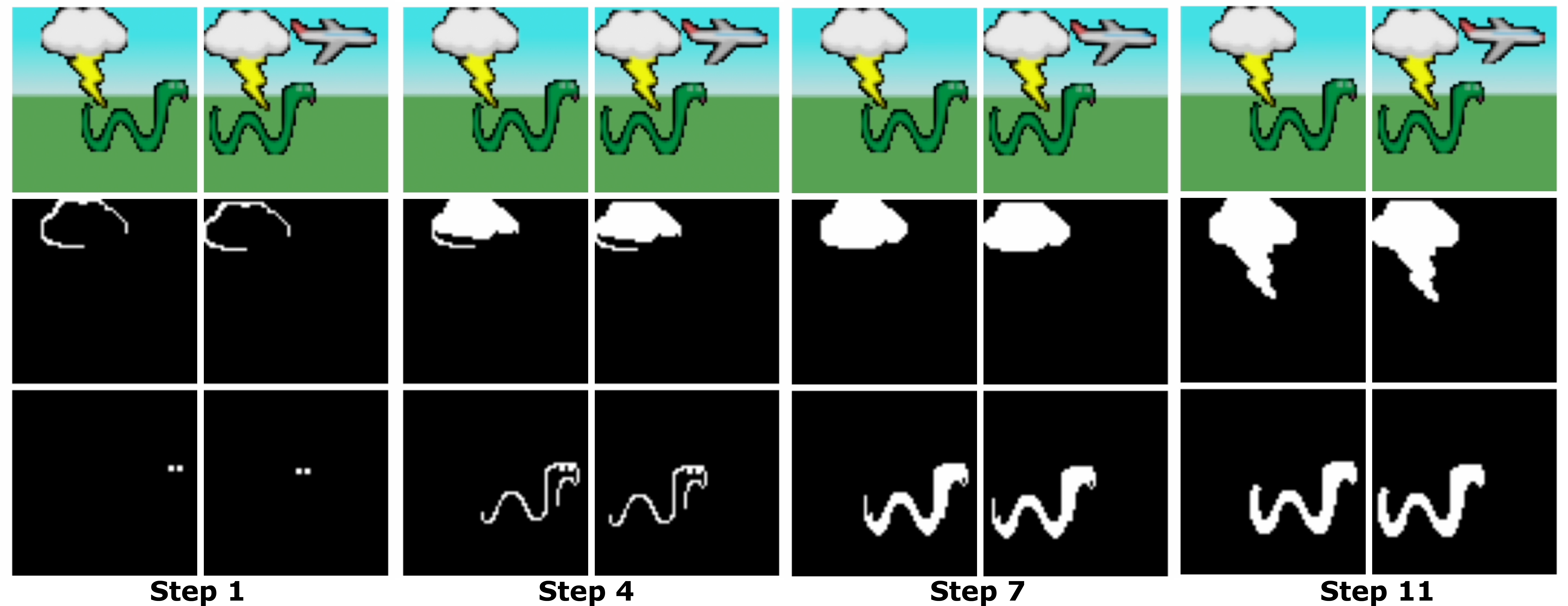} \label{fig:res_objdisc_demo_1}}
  \caption{Supplementary visualization of Co-Part Object Discovery Algorithm.} \label{fig:res_objdisc_demo}
\end{figure*}

In Figure \ref{fig:res_objdisc_demo}, we visualize the runtime merging of parts into objects using our co-part object discovery algorithm  with samples from AbsScene.

\subsection{Object Discovery}
\label{appn:visuals-obj-disc}
\begin{figure*}[t!]
  \centering
  \subfloat[Tetrominoes]{\includegraphics[width=0.99\linewidth]{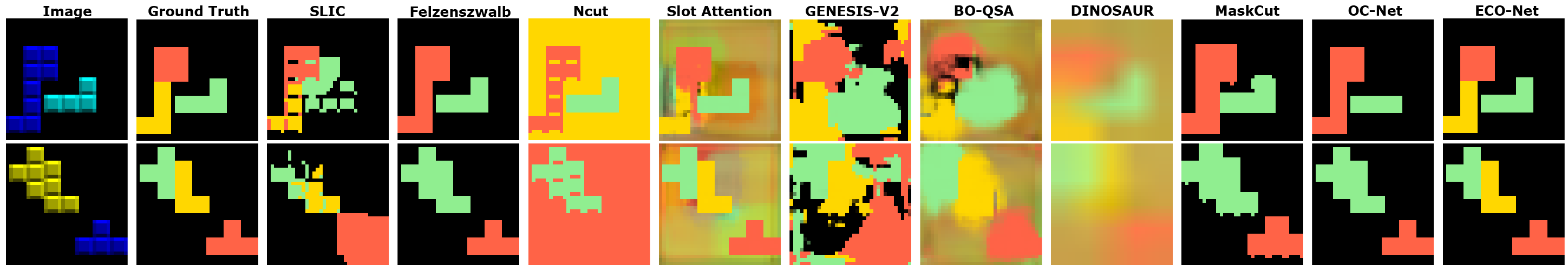} \label{fig:res_objdisc_sim_te_supp}}\\
  \subfloat[AbsScene]{\includegraphics[width=0.99\linewidth]{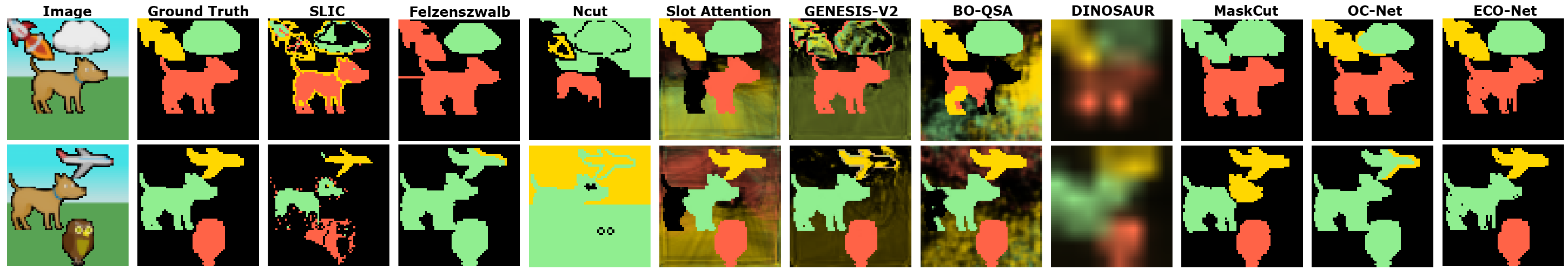} \label{fig:res_objdisc_sim_absscene_supp}}\\
  \caption{Supplementary visualization of discovered objects on simulated datasets.} \label{fig:res_objdisc_sim_supp}
\end{figure*}

\begin{figure*}[t!]
  \centering
  \subfloat[GSO]{\includegraphics[width=0.99\linewidth]{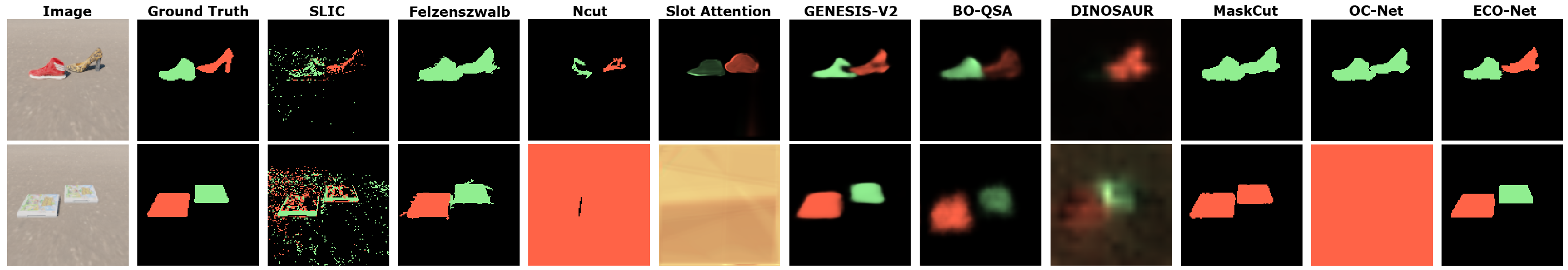} \label{fig:res_objdisc_gso_supp}}\\
  \subfloat[SKU-110K]{\includegraphics[width=0.99\linewidth]{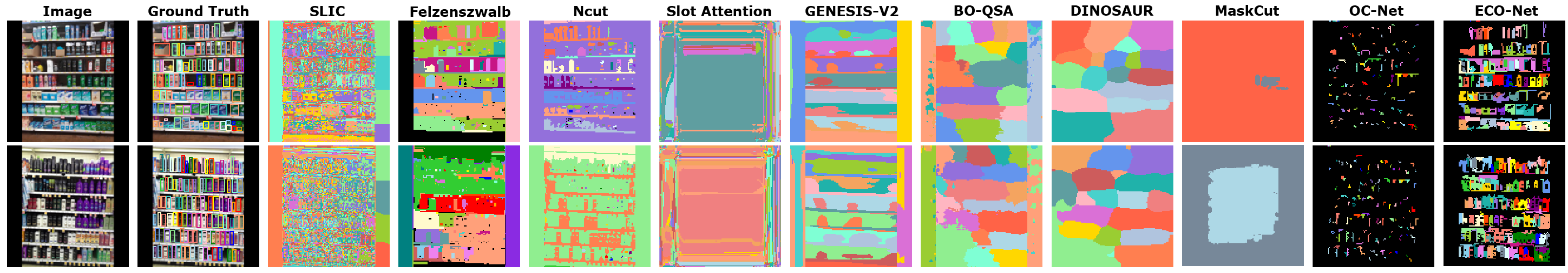} \label{fig:res_objdisc_sku110k_supp}}
  \caption{Supplementary visualization of discovered objects on realistic datasets.} \label{fig:res_objdisc_real_supp}
\end{figure*}

\begin{figure*}[t!]
  \centering
  \subfloat[PASCAL VOC]{\includegraphics[width=0.99\linewidth]{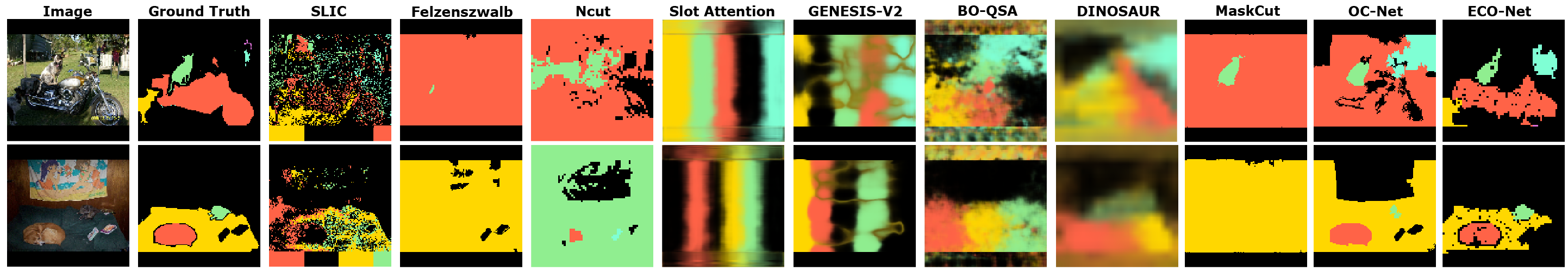} \label{fig:res_objdisc_pascal_supp}}\\
  \subfloat[MS COCO]{\includegraphics[width=0.99\linewidth]{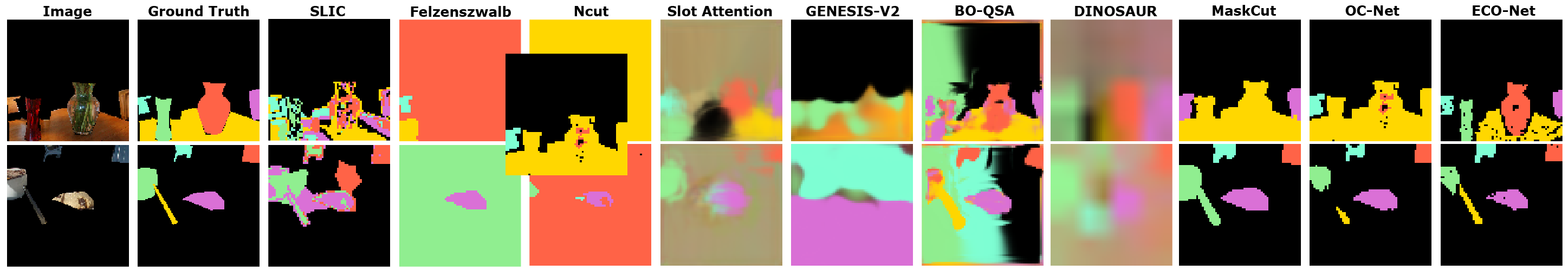} \label{fig:res_objdisc_coco_supp}}
  \caption{Supplementary visualization of discovered objects on real-world datasets.} \label{fig:res_objdisc_realworld_supp}
\end{figure*}

We visualize the results of additional samples from simulated datasets in Figure \ref{fig:res_objdisc_sim_supp}, realistic datasets in Figure \ref{fig:res_objdisc_real_supp} and real-world datasets in Figure \ref{fig:res_objdisc_realworld_supp}.

% \clearpage
\subsection{Occlusion-Aware Perception}
\label{appn:visuals-occlu}
\begin{figure*}[h!]
  \centering
  \subfloat[AbsScene-O]{\includegraphics[width=0.99\linewidth]{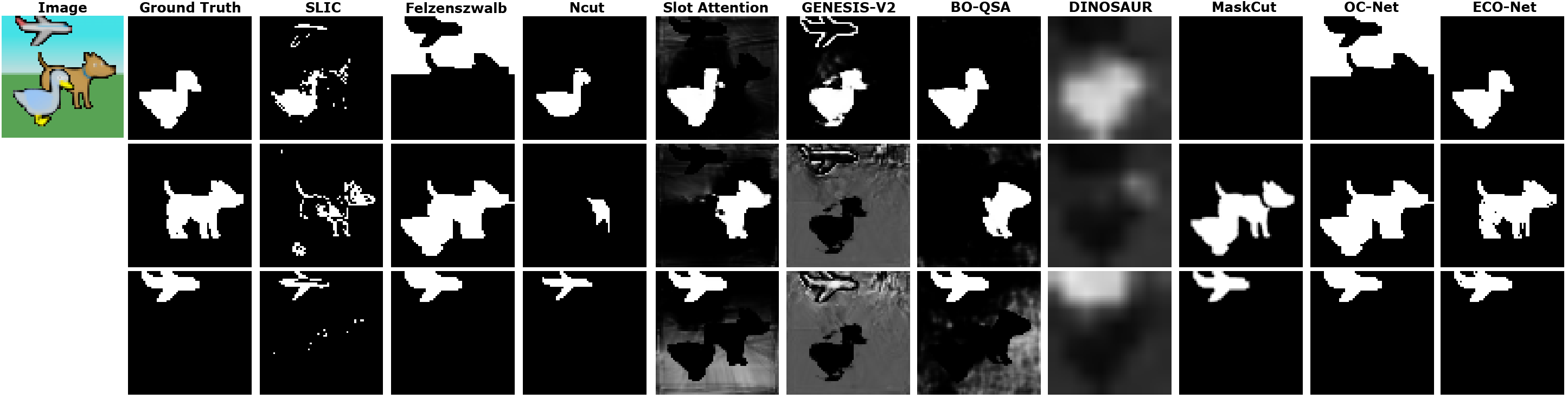} \label{fig:res_occlu_sim_absscene_supp}}\\
  \subfloat[GSO-O]{\includegraphics[width=0.99\linewidth]{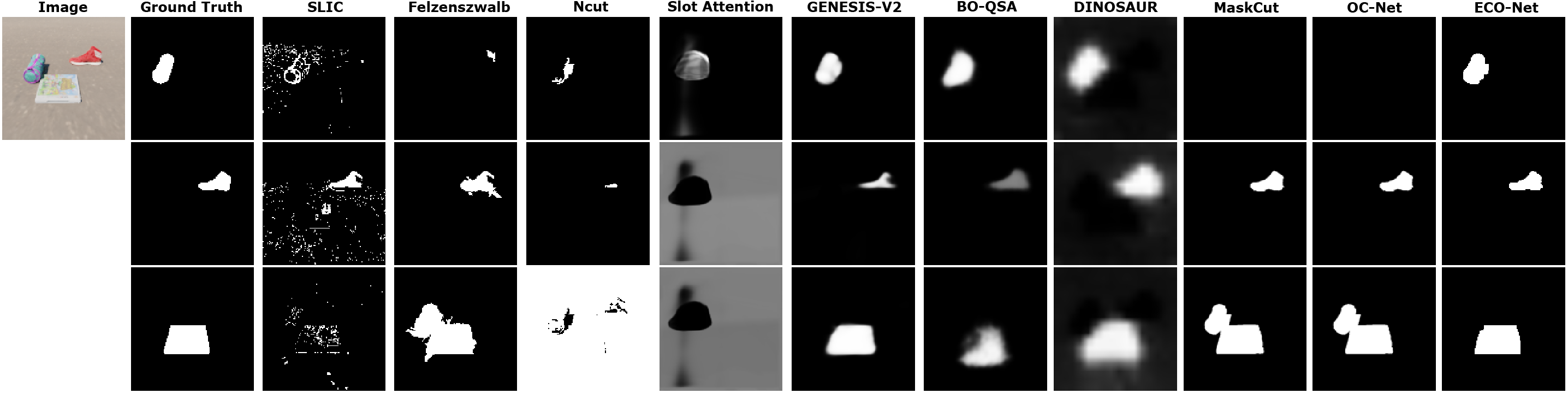} \label{fig:res_occlu_real_gso_supp}}
  \caption{Supplementary visualization of discovered objects on occluded objects datasets.} \label{fig:res_objdisc_occlu_supp}
\end{figure*}

We visualize the results of additional samples from occluded-objects datasets in Figure \ref{fig:res_objdisc_occlu_supp}.

% \clearpage
\subsection{Out-of-Distribution Object Discovery}
\label{appn:visuals-ood}
\begin{figure*}[h!]
  \centering
  \includegraphics[width=0.99\linewidth]{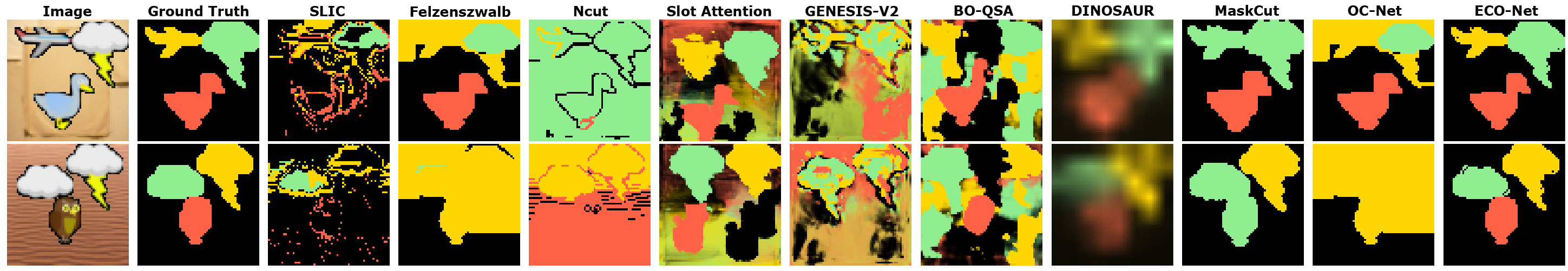} 
  \caption{Supplementary visualization of discovered objects on out-of-distribution test set AbsScene-C.}\label{fig:res_gen_supp}
\end{figure*}

We visualize the results of additional samples from AbsScene-C in Figure \ref{fig:res_gen_supp}.

\section{Limitations}
\label{limitations}
There are still obstacles that have to be overcome for successful application of our framework to the visual complexity of the real world. 
A natural next step would be to extend ECO-Net to more efficient implementations of the co-part object discovery algorithm especially for applications which require real-time object discovery.
It is also promising to dynamically learn an appropriate clustering similarity constant given the training data.
Lastly, real-world scenes with multi-part objects is still of higher visual complexity than the datasets considered here and reliably bridging this gap is an open problem. 

\section{Ethics Statement}
\label{ethics}
Our analysis, which is focused on publicly available multi-object simulated and real-world data has no immediate impact on general society. To the best of our knowledge, none of the datasets used in this study contain personally identifiable information or offensive content. As with any model that performs scene understanding, applications with potential negative societal impact such as in the area of surveillance cannot be fully excluded upon future research in this area. At this point in time, however, there remains challenging open questions that need to be reliably addressed before enabling the deployment of ECO-Net and its counterparts to real-world settings, and hence a direct application of this method for malicious purposes is currently unlikely.

\end{document}